\documentclass{article}
\usepackage{ijcai25}

\usepackage{times}
\usepackage{soul}
\usepackage{url}
\usepackage[hidelinks]{hyperref}
\usepackage[utf8]{inputenc}
\usepackage[small]{caption}
\usepackage{graphicx}
\usepackage{amsmath}
\usepackage{amsthm}
\usepackage{booktabs}
\usepackage{algorithm}
\usepackage{algorithmic}
\usepackage[switch]{lineno}

\usepackage{subcaption}
\usepackage[dvipsnames]{xcolor}

\title{TRIP: A Nonparametric Test to Diagnose Biased Feature Importance Scores}

\author{Aaron Foote$^1$ \and Danny Krizanc$^1$\\
\affiliations $^1$Wesleyan University\\
\emails \{afoote,dkrizanc\}@wesleyan.edu
}

\begin{document}

\maketitle

\begin{abstract}
Along with accurate prediction, understanding the contribution of each feature to the making of the prediction, i.e., the importance of the feature, is a desirable and arguably necessary component of a machine learning model. For a complex model such as a random forest, such importances are not innate – as they are, e.g., with linear regression. Efficient methods have been created to provide such capabilities, with one of the most popular among them being permutation feature importance due to its efficiency, model-agnostic nature, and perceived intuitiveness. However, permutation feature importance has been shown to be misleading in the presence of dependent features as a result of the creation of unrealistic observations when permuting the dependent features. In this work, we develop TRIP (\textbf{T}est for \textbf{R}eliable \textbf{I}nterpretation via \textbf{P}ermutation), a test requiring minimal assumptions that is able to detect unreliable permutation feature importance scores that are the result of model extrapolation. To build on this, we demonstrate how the test can be complemented in order to allow its use in high dimensional settings. Through testing on simulated data and applications, our results show that the test can be used to reliably detect when permutation feature importance scores are unreliable.
\end{abstract}

\section{Introduction}

For structured tabular data, tree-based models have shown performance comparable to or better than deep learning approaches \cite{forestOverDL1,forestOverDL2}. For this work, we focus on the random forest method \cite{randomForest}, a tree-ensembling approach that utilizes bagging and random feature selection to construct a sufficiently diverse ensemble, typically yielding good performance with minimal tuning \cite{minimalTuning}.

A key consideration along with performance is the interpretability of the model, i.e., an understanding of how predictions are produced from input data. It is a necessary part of safely deploying machine learning models in many settings \cite{needInterpretability1}, and it is an important attribute of models that hope to see widespread deployment in settings where high-stakes decisions are made. Random forests are not naturally interpretable, as they employ an bagged ensembling approach. With their popularity, there has been much work on improving their interpretability. One of the most widely used feature importance methods are split-improvement measures \cite{CART}. However, it has been observed that this approach gives increased importance to features that have many possible splits, such as continuous features or features with many categories \cite{giniBiased}. With this, a slew of alternative methods have been proposed, such as permutation feature importance (PFI) \cite{randomForest}, partial dependence plots (PDP) \cite{partialDependencePlot}, individual conditional expectation plots (ICE) \cite{individualConditionalExpectationPlot}, and the more widely used SHAP method \cite{SHAP}. There has been other work on diagnosing the reliability of outputs for tree-based models in concept drift/distribution shift as well \cite{conceptDrift}.

The methods of PDP, ICE, and PFI are all widely used for auditing machine learning models due to their ease of use, efficiency, and perceived intuitiveness. Another element common to all of these methods is that they create new data points upon which the model is evaluated. If the features of the data are independent -- a strong assumption -- the models produce useful results. In the presence of correlated or dependent features, the newly generated points can deviate considerably from the training distribution, potentially being entirely unrealistic \cite{noFreeVariableImportance,functionalANOVA}. Along with a more precise characterization and explanation \cite{noFreeVariableImportance}, this phenomenon has been demonstrated through simulation \cite{ARCHER,conditionalPermutation}.

As an accompanying model, the conditional permutation feature importance (cPFI) scheme includes a conditioning step before permuting \cite{strobl3,revisitCPFI}. This approach has proved effective for small datasets ($n \leq 500$), but becomes computationally intractable as the sample size increases \cite{conditionalPermutation}. Another more robust approach to evaluating importance is to retrain the model after removing the effect of a feature, as this does not risk model performance being evaluated on a test set that differs considerably from the training set. This involves permuting and relearning \cite{permRelearn} or dropping and relearning \cite{loco}. Simulations suggest that these approaches do not fall prey to the extrapolation bias of unmodified PFI \cite{noFreeVariableImportance}. However, training the model over and over on datasets that are only one feature smaller can be prohibitively costly.

In this paper, we propose a nonparametric hypothesis testing framework by which the reliability of vanilla PFI scores can be evaluated, using the extent to which the model extrapolates upon a feature's permuting as a measure of reliability of the importances. With this, a user does not need to jump to computationally intensive methods straight away, instead assessing the validity of the efficiently computed feature importances to decide whether the computationally intensive approaches are necessary. In Section 2 we develop the method, testing it on simulated data and applications in Section 3. In Section 4, we discuss the performance of the method in higher dimensions, offer an additional step that can be done to ensure reliable performance, and demonstrate the improvement with applications to simulated data and gene expression data sets. Section 5 presents our conclusions.

\section{Proposed Method}
In order to identify unreliable PFI scores, we will measure the extrapolation that occurs when a feature is permuted. This circumvents the difficult task of computing dependencies between features, and cuts straight to the issue that causes unreliable feature importance scores. Note that in a decision tree (and by extension random forest), for any input point $\mathbf{x} = (x_1,\dots,x_m)$, the only training points that are used to generate the prediction are those in the leaf in which $\mathbf{x}$ ends up. We term this set of points the \emph{leaf community} of $\mathbf{x}$, written $\mathcal{L}(\mathbf{x})$. Then, the extrapolation of a model will be measured by the average distance from an input point to its leaf community. For a random forest with $t$ trees, we simply take the average of all of the trees: \[\frac{\sum_{i=1}^t \sum_{\mathbf{x}_\ell \in \mathcal{L}_i(\mathbf{x})} d(\mathbf{x},\mathbf{x}_\ell)}{\sum_{i=1}^t \vert\mathcal{L}_i(\mathbf{x})\vert}.\] Now, each time a feature is permuted to calculate the increase in error, the average leaf community distance (ALCD) can be computed for each point, yielding a measure of extrapolation. However, even when permuting a feature that is independent, we would expect to see input points have non-zero average leaf community distance. It would then be difficult to discern whether the smaller values were close to independent or just the least dependent among the features. To account for this, a baseline feature, sampled uniformly on the zero-one interval, is added to the model before training. This feature is guaranteed to be independent, and thus we can compare ALCDs for all of the other features to the baseline.

Rather than just eyeballing distances however, we propose a nonparametric permutation testing approach to investigate extrapolation. The data we are working with is the ALCD for observations resulting from permuting each feature, including the baseline. If there is no extrapolation, the ALCD for each feature should be equal to that of the ALCD resulting from permuting the baseline feature, on average.  

The key assumption for the permutation test is exchangeability, i.e., that under the null hypothesis -- the average paired difference is zero -- group labels can be exchanged without modifying the joint probability distribution of the labeling. This clearly holds, as under the null hypothesis ALCD from permuting a feature should be equal to the ALCD from permuting the baseline, so the labels can be exchanged.

For each feature, we compute our test statistic of the average of differences in average leaf community distances upon permuting. Then, given exchangeability, we are free to shuffle the group labels by randomly assigning group labels within each pair, doing this permuting repeatedly and calculating a new statistic each time (Alg.~\ref{alg:algorithm}). This shuffling and recomputing process is repeated for a set number of permutations, at which point we have a distribution from which we can calculate our p-value.

\begin{algorithm}[bt]
    \caption{Permutation Test for Extrapolation}
    \label{alg:algorithm}
    \textbf{Input}: ALCD for each Feature\\
    \textbf{Parameter}: Permutation Count $\pi$\\
    \textbf{Output}: p-value for each feature
    \begin{algorithmic}[1] 
        \STATE $pVals$ = empty array
        \FOR{\textbf{each} $j$ in $\{1,\dots,p\}$}
            \STATE $ALCDs^{(j)} = averageDistances[j]$
            \STATE $s = \frac{1}{m}\sum_{i=1}^m ALCDs^{(j)}_i - ALCDs^{(B)}_i$
            \STATE Let $permIdx=0$.
            \WHILE{$permIdx < \pi$}
                \STATE Randomly assign group labels within pairs, yielding $ALCDs^{(j,\pi)}$, $ALCDs^{(B,\pi)}$
                \STATE New test statistic $= \frac{1}{m}\sum_{i=1}^m ALCDs^{(j,\pi)}_i - ALCDs^{(B,\pi)}_i$
            \ENDWHILE
            \STATE let $tail$ = \# of test statistics more extreme than $s$
    	\STATE $pVals[j]$ = $tail/\pi$
        \ENDFOR
    \end{algorithmic}
\end{algorithm}

\section{Simulated Data and Applications}
We now explore the ability for the test to detect unreliable PFI scores due to extrapolation upon permutation of dependent features by applying it to simulated and real data. The computationally intensive cPFI method will be used as the ground truth importance. In the presence of feature dependence and extrapolation upon permutation, PFI should be considerably greater than cPFI, and the p-values of the test concentrated lower than for independent features.

\subsection{Setup}
To start, the method will be applied to four simulated data sets. The first, similar to the one used to explain the extrapolation phenomenon \cite{noFreeVariableImportance}, generates the data as a linear model \[y_i = \varepsilon_i + \sum_{j=1}^8 x_{ij},\] where the first two features come from a multivariate normal distribution with means zero and covariance matrix \[\begin{bmatrix}
    1       & \rho \\
    \rho    & 1     
\end{bmatrix}.\] The remaining six features are sampled uniformly on the zero-one interval, making them independent, with the noise $\epsilon \sim N(0,0.1^2)$. The second keeps the last six features the same, but instead generates the first two as
\begin{align*}
    \theta  &\sim Unif(0,2\pi) &  x_1     &= sin(\theta) & x_2     &= cos(\theta).
\end{align*} With this, the first two features are uncorrelated, yet are still dependent upon one another. We would also like to consider situations where most of the features are dependent. For this, we will generate all features from a multivariate normal distribution with means of zero and the covariance matrix taking on a block structure in which groups of $k$ features are correlated with each other but independent of the rest of the features. If the block size $k$ does not divide the number of features $p$, then the leftover features will be left as independent.

A correlation of $\rho = 0.75$, block size of three, sample size of 500 is used. We allow blocks to have different correlations as well, and for this we take $\rho_1 = 0.75$ and $\rho_2 = 0.25$. Each data generating process is run ten times, and when permuting to calculate feature importances/distances we permute 25 times. On each iteration, after generating the data and fitting the forest, both vanilla PFI and cPFI are computed. The implementation is provided by the \texttt{permimp} package in R \cite{cPFI}. Next, we will be using the wine classification dataset publicly available in the UCI Machine Learning Repository \cite{Wine}. It has 178 observations and 13 features capturing properties of the wine, all of which was grown in the same region of Italy. The goal is to classify the wines by cultivar, of which there are three. Using 75\% of the data for training, an untuned random forest was trained using the \texttt{randomForest} package \cite{randomForestPackageR}. 

\subsection{Results}
Plotted are the permutation feature importance scores (conditional in \textcolor{OliveGreen}{green} and unconditional in \textcolor{RedOrange}{orange}) on the left hand side, with a boxplot of average importances over 25 permutations, for the 10 runs. On the right is a boxplot of p-values for each feature, with each box containing 250 p-values from each permutation of each data generation. The features that we know to be dependent have \textcolor{red}{red} boxes, and the independent ones \textcolor{Blue}{blue}.

\begin{figure}[htbp]
	\centering
	\captionsetup{justification=centering}
	\subfloat[Feature importances, PFI in \textcolor{RedOrange}{orange} and cPFI in  \textcolor{OliveGreen}{green}\label{fig:twoCorrelatedImps}]{\includegraphics[width=\columnwidth]{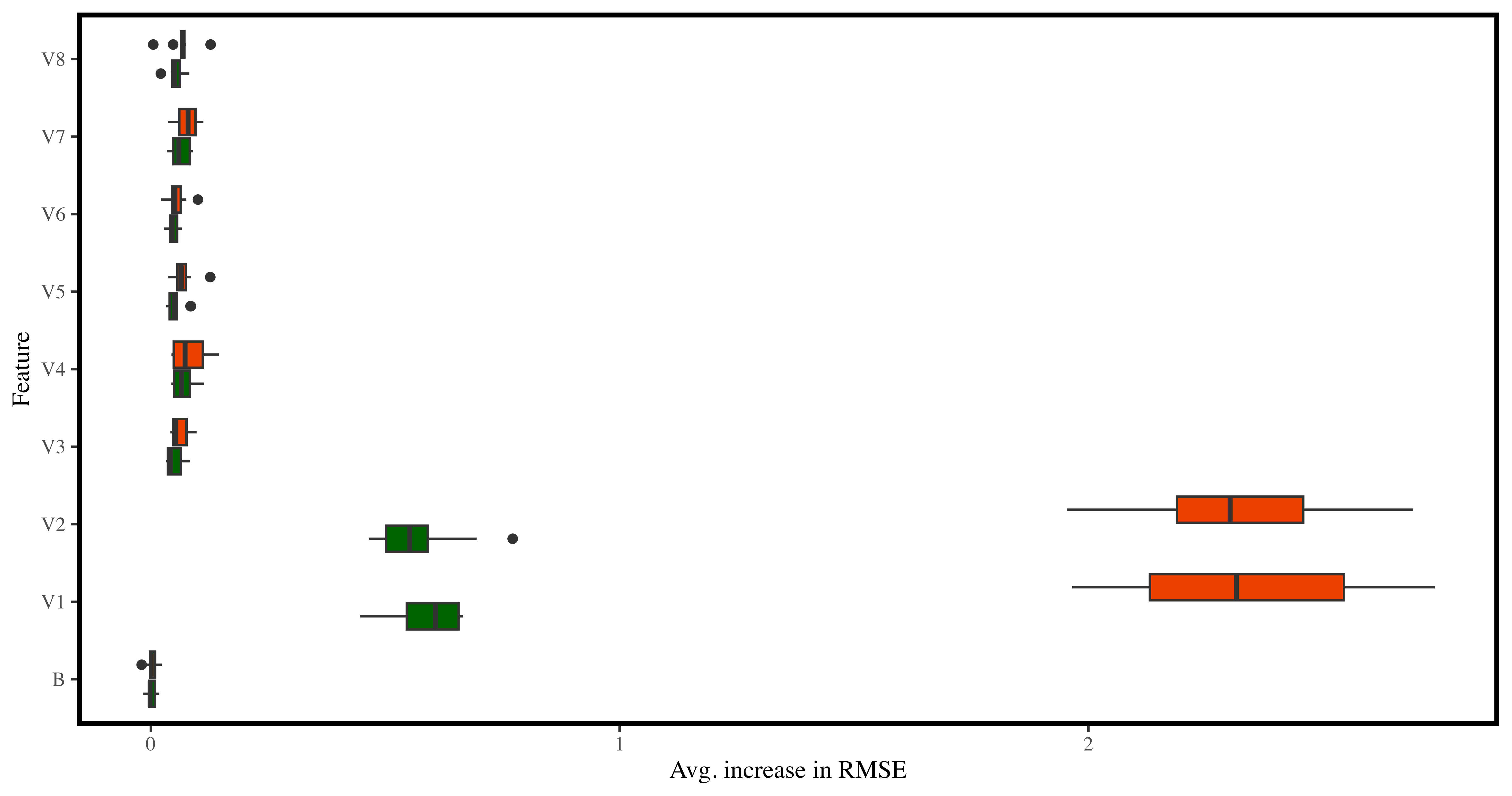}}\hfil
	\subfloat[Boxplots of p-values for each feature.\label{fig:twoCorrelatedPVals}]{\includegraphics[width=\columnwidth]{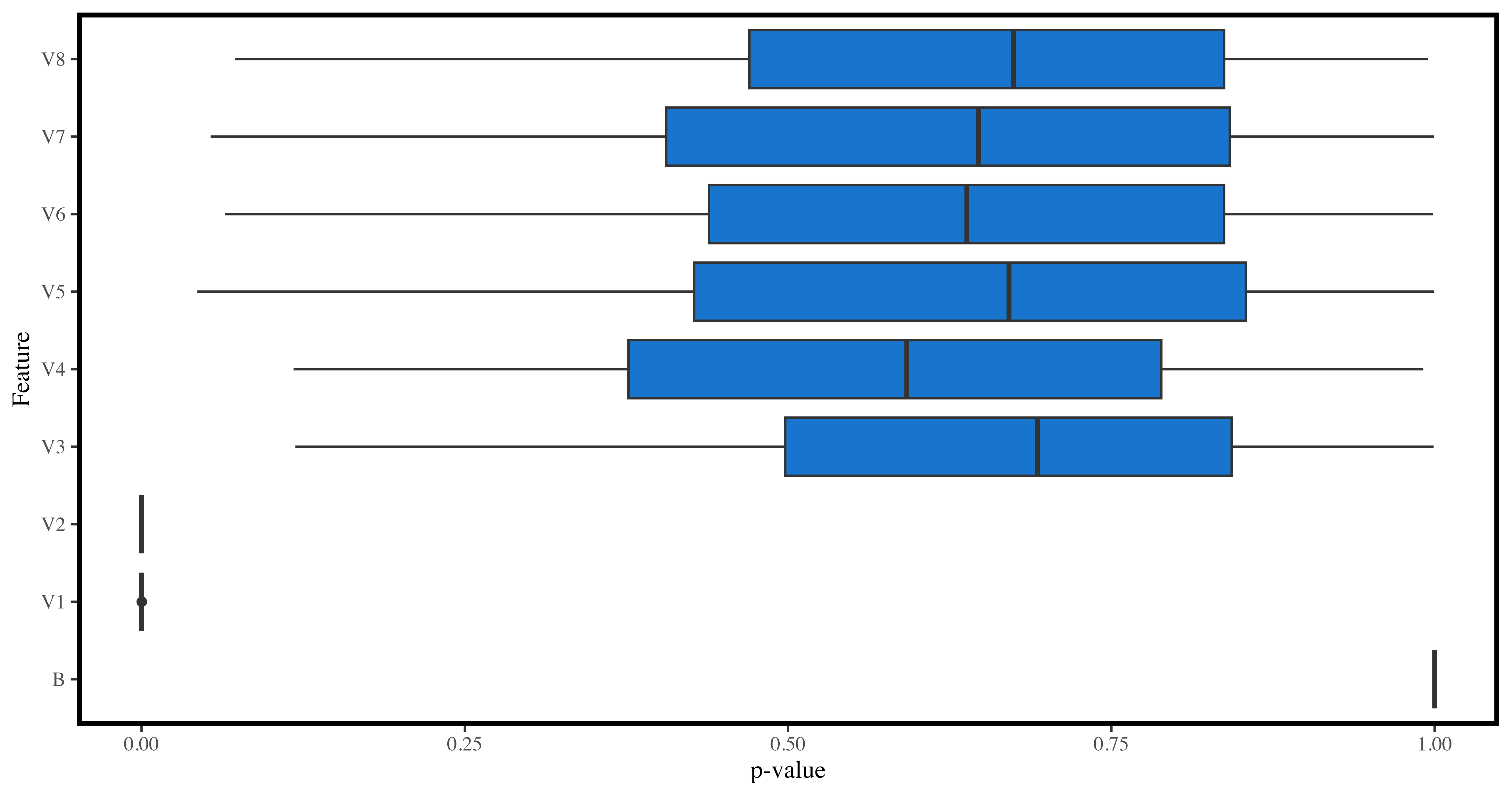}}
	\caption{Feature importance/testing for the simulated dataset. Data generated with \textbf{two correlated features, the rest independent}.}
	\label{fig:simApply2c}
\end{figure}

For the generating process including two correlated features (Fig. \ref{fig:simApply2c}), we see that the two correlated features have PFI scores much greater than cPFI scores, indicating that the importance scores are inflated. This is captured by the test, which produces boxplots of p-values for these two features taking values between 0.0001 and 0.0002. Additionally, for the rest of the features, which are each independent of all other features, the median p-value is well above 0.5 and the two importance scores report comparable feature importances. Note that the increased importance of the two correlated features makes sense, as they do in fact provide additional information for learning, leading to a higher cPFI score. However, the additional importance they carry is not as much as PFI would lead one to believe. 

\begin{figure*}[tb]
	\centering
	\captionsetup{justification=centering}
	\subfloat[Feature importances, PFI in \textcolor{RedOrange}{orange} and cPFI in \textcolor{OliveGreen}{green}\label{fig:justDependentImps}]{\includegraphics[width=\columnwidth]{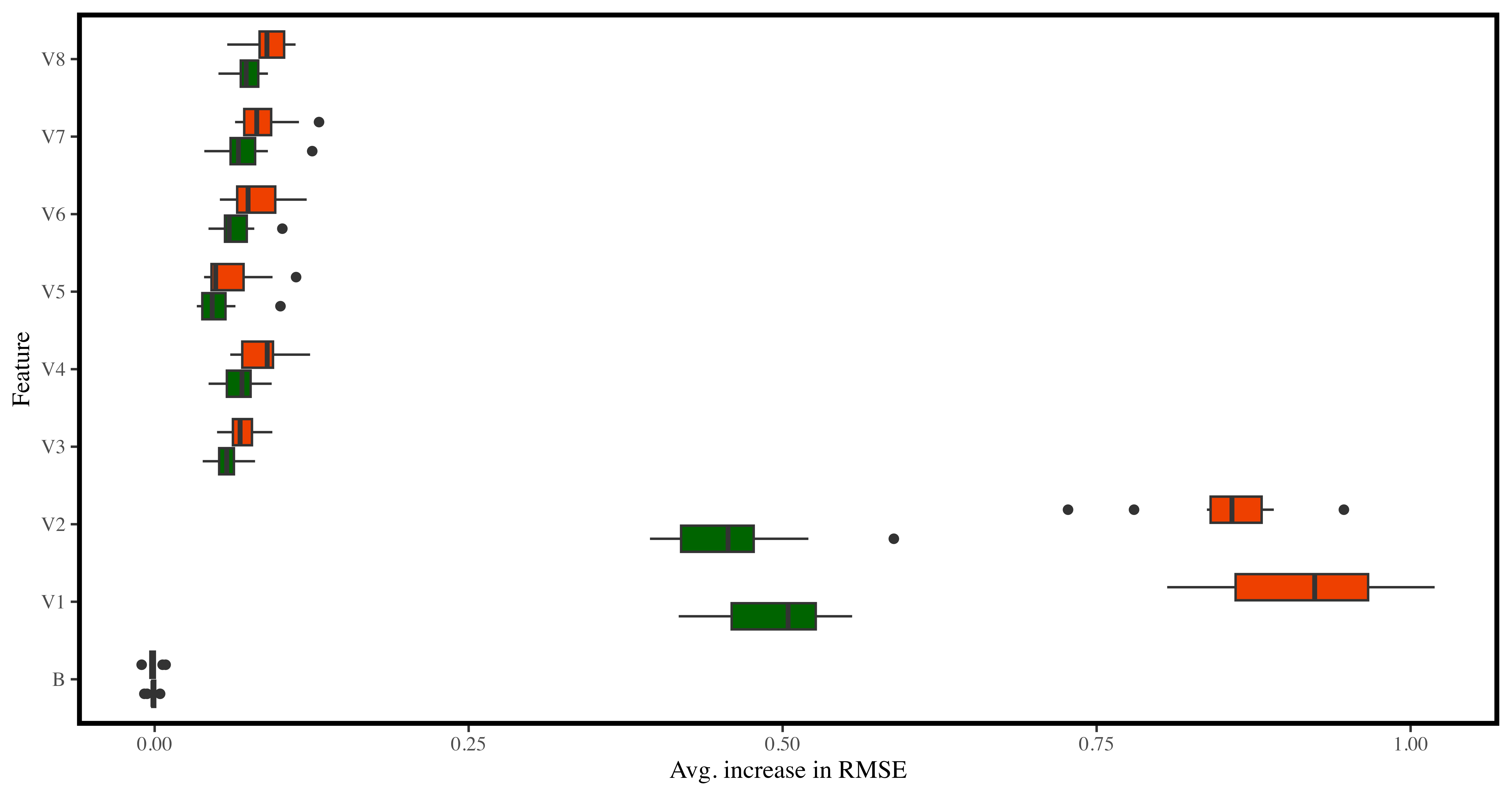}}\hfil
	\subfloat[Boxplots of p-values for each feature.\label{fig:justDependentPVals}]{\includegraphics[width=\columnwidth]{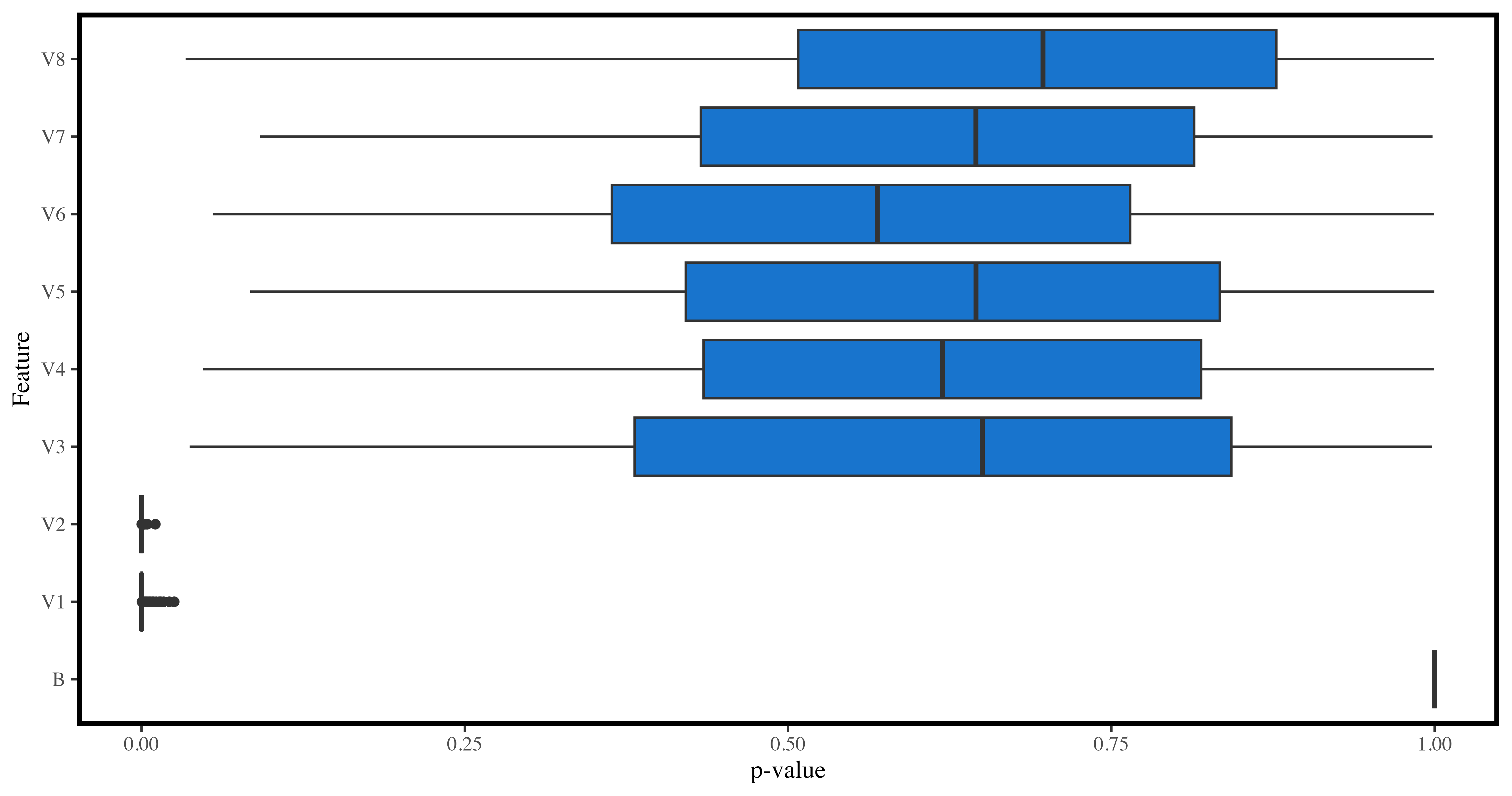}}
	\caption{Feature importance/testing for the simulated dataset. Data generated with \textbf{two dependent but uncorrelated features, the rest independent}.}
	\label{fig:simApply2d}
\end{figure*}

We see similar results for the generating process including two dependent features (Fig. \ref{fig:simApply2d}). The two dependent features have much greater PFI scores than they do cPFI scores, indicating that their importances have been inflated by extrapolation upon permutation. Note here that with these two features being perfectly dependent upon one another, they are in fact more useful in making predictions than the uniform random features that are V3 through V8. This generating process shows that the test is not limited just to correlations, which makes sense, as correlations were not used in the calculation of distances or in permuting.

\begin{figure*}[tb]
	\centering
	\captionsetup{justification=centering}
	\subfloat[Feature importances, PFI in \textcolor{RedOrange}{orange} and cPFI in  \textcolor{OliveGreen}{green}\label{fig:blocksImps}]{\includegraphics[width=\columnwidth]{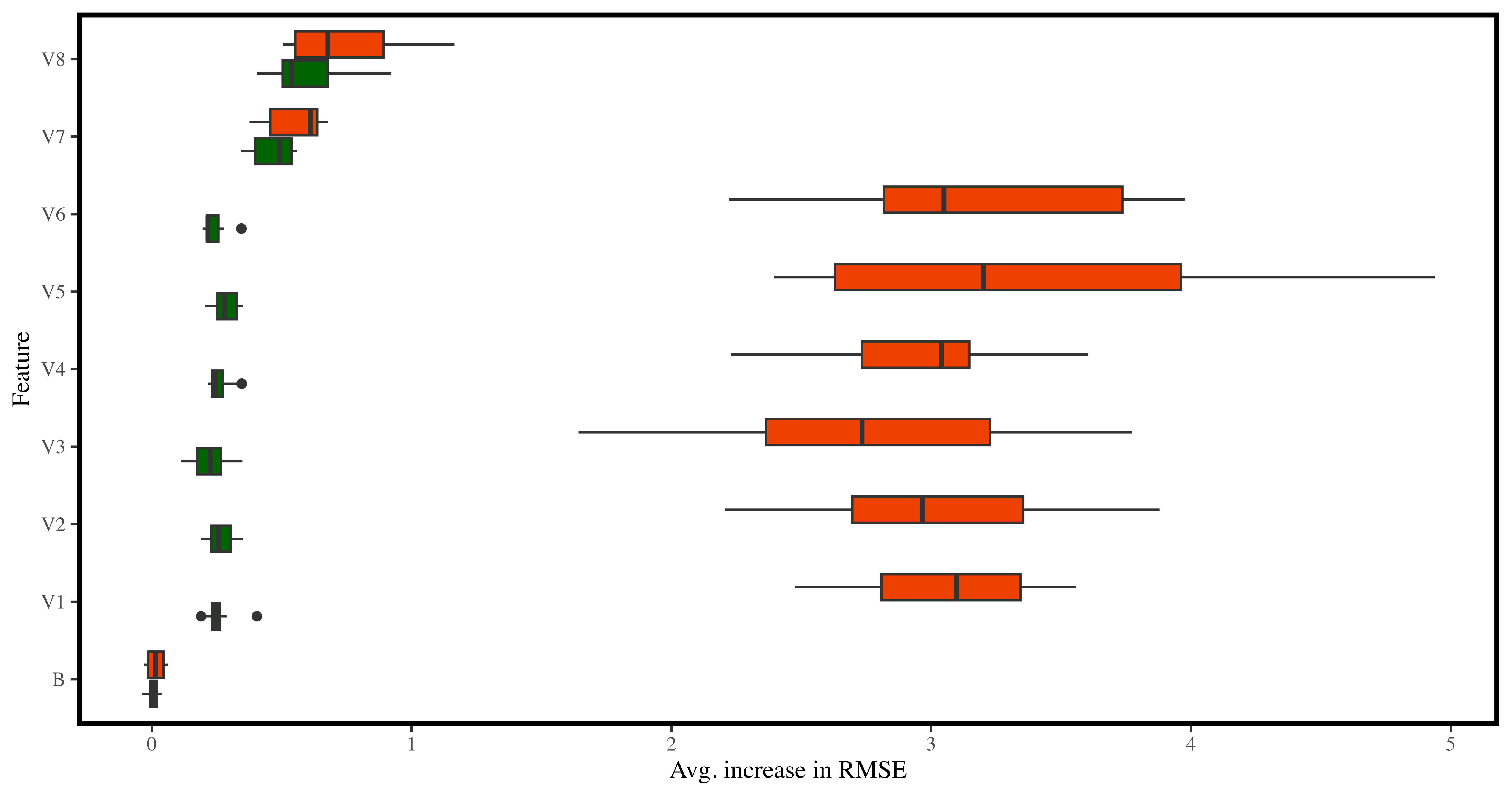}}\hfil
	\subfloat[Boxplots of p-values for each feature.\label{fig:blocksPVals}]{\includegraphics[width=\columnwidth]{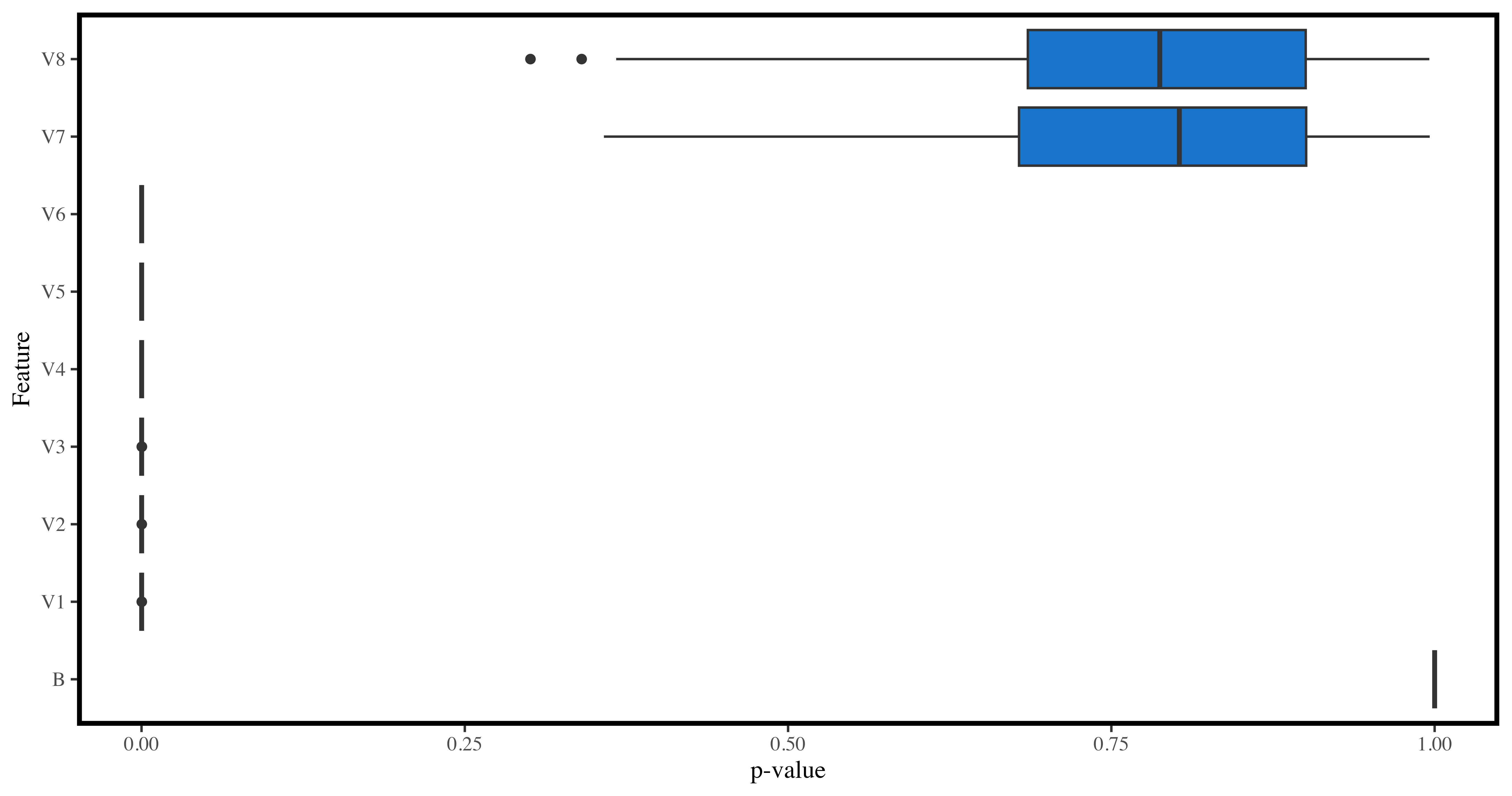}}
	\caption{Feature importance/testing for the simulated dataset. Data generated with \textbf{blocks of correlated features with equal correlation}.}
	\label{fig:simApplyBlocks}
\end{figure*}

\begin{figure*}[tbh]
	\centering
	\captionsetup{justification=centering}
	\subfloat[Feature importances, PFI in \textcolor{RedOrange}{orange} and cPFI in  \textcolor{OliveGreen}{green}\label{fig:meddlesomeImps}]{\includegraphics[width=\columnwidth]{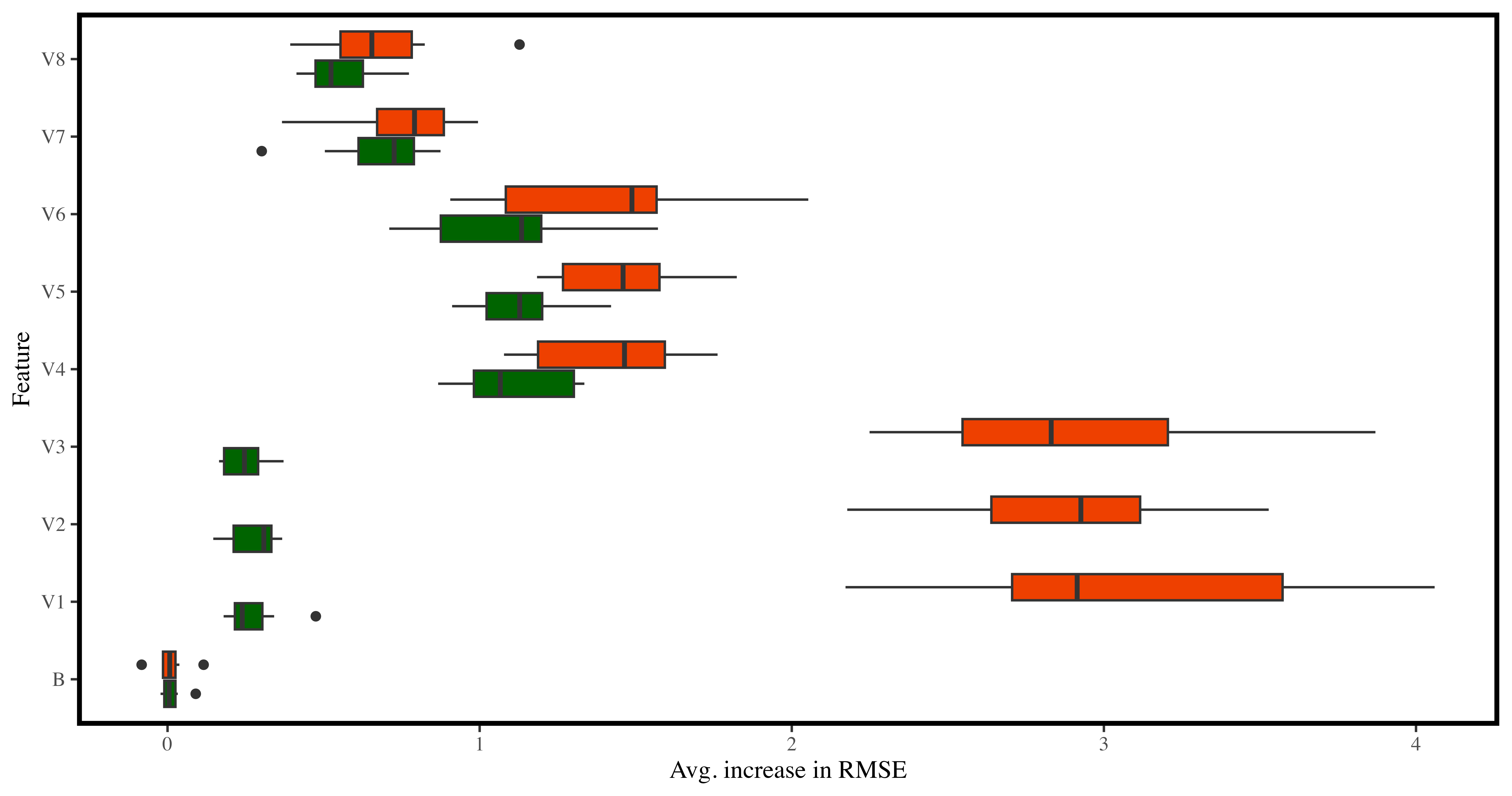}}\hfil
	\subfloat[Boxplots of p-values for each feature.\label{fig:meddlesomePVals}]{\includegraphics[width=\columnwidth]{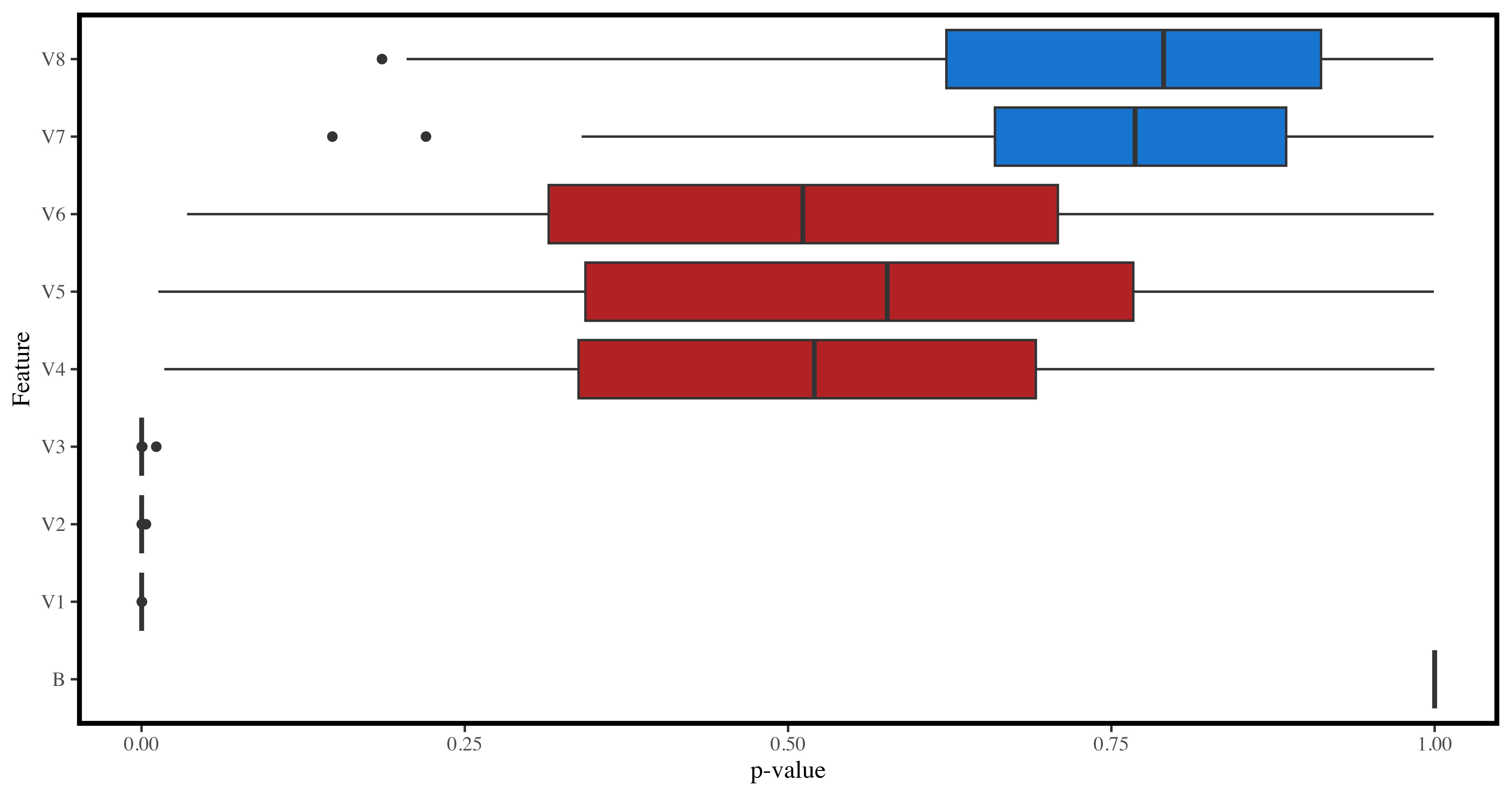}}
	\caption{Feature importance/testing for the simulated dataset. Data generated with \textbf{blocks of correlated features with varied correlation}.}
	\label{fig:simApplyMeddlesome}
\end{figure*}

The true power of the test can be seen for the two generating processes that make most of the features dependent. For the blocks of equal correlation (Fig. \ref{fig:simApplyBlocks}), PFI overestimates the importance of the correlated features (V1 through V6) considerably, which is also reflected in the p-values for these features. In fact, they are overestimated to the point that PFI would rank the independent features (V7 and V8) as the two least important ones (ignoring the baseline feature that we know is not useful) while in reality these are the most important features by cPFI. For the two independent features, both feature importance methods provide similar scores, and the p-values having a median above 0.75 reflect this lack of extrapolation. For the blocks with varied correlation (Fig. \ref{fig:simApplyMeddlesome}), the first three features V1, V2, and V3 have correlation of $\rho = 0.75$ with one another and the next three features V4, V5, and V6 have correlation $\rho = 0.25$ with one another, leaving the last two as independent. We see that for the first block, we get the same behavior as before, with inflated PFI scores and p-values with lower and upper quantiles of 0.0001. However, for the next block, the weaker correlation leads to less extrapolation, as can be seen by the minimally inflated PFI scores relative to the cPFI scores. In the plot of p-values, this behavior is reflected by p-values with lower quantiles between 0.315 and 0.343 and upper quantiles between 0.692 and 0.767 which are not as high as for the independent features, but are up to the practitioner to interpret and less clearly cut. It turns out that the first block of features are improperly ranked by PFI, but this minor inflation due to minor correlation is not enough to create incorrect rankings if PFI is used for the second block. This is certainly not guaranteed however, and in some settings a practitioner may decide to try out the more computationally intensive cPFI just to make sure that their rankings are accurate. We emphasize that a traditional threshold of 0.05 may not be the best way to determine if there is extrapolation, as in certain settings where many features are closely ranked, p-values above 0.05 may still push a user to use more computationally intensive approaches.

In the case of the wine classification data, for the features with the greatest disparity in PFI and conditional PFI scores (proline, flavonoids, color intensity, and alcohol), we see the boxplots of p-values cover a much lower range of p-values than the features for which PFI does not greatly overestimate the importance (proanthocyanins, malicacid, and ash). Another characteristic of the boxplots for features with large disparities in PFI and conditional PFI scores is a much smaller spread in p-values. Keep in mind that without computing conditional importance scores, the user would only have the orange bars the plot on the right. With the boxplots suggesting that the most important features are exhibiting extrapolation, a user could opt for the conditional PFI approach, obtaining more accurate importances.

\begin{figure*}[tb]
	\centering
	\captionsetup{justification=centering}
	\subfloat[Feature importances, PFI in \textcolor{RedOrange}{orange} and cPFI in  \textcolor{OliveGreen}{green}\label{fig:wineImps}]{\includegraphics[width=\columnwidth]{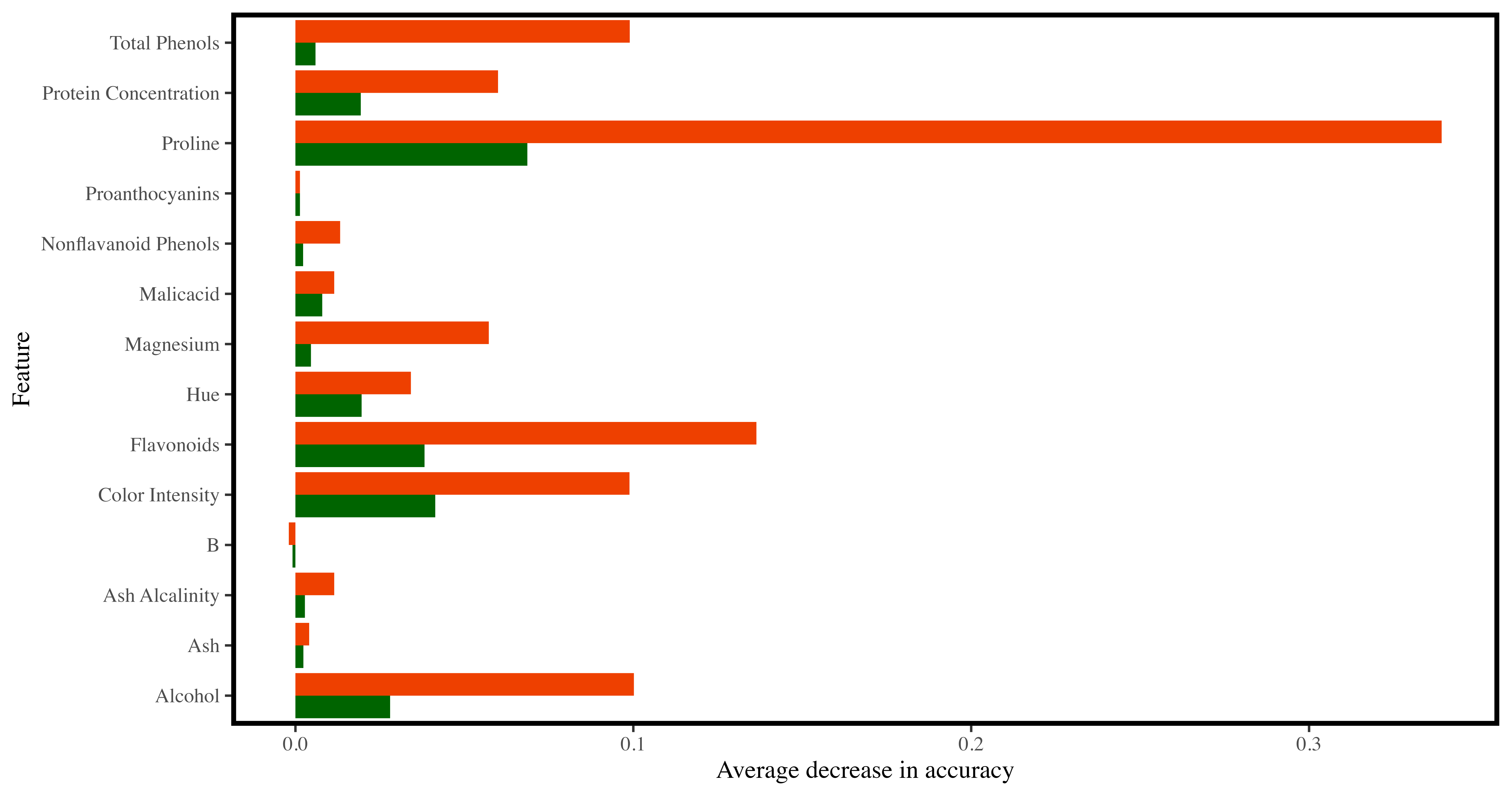}}
	\subfloat[Boxplots of p-values for each feature\label{fig:penguinPVals}]{\includegraphics[width=\columnwidth]{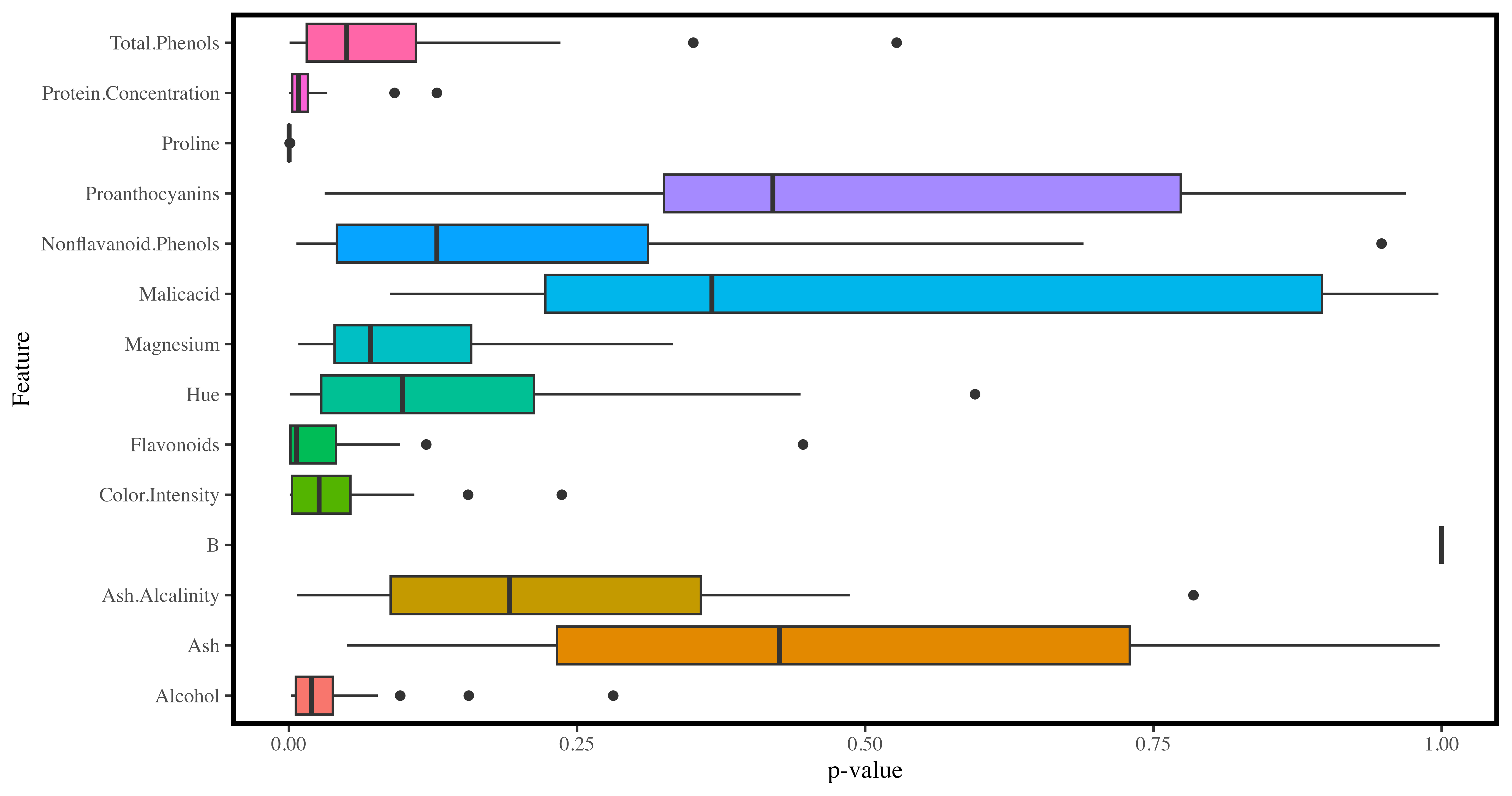}}
	\caption{Feature importance/testing the wine classification dataset. The PFI scores for most features are overestimated, which is reflected in the boxplots of p-values on the right-hand plot.}
	\label{fig:wineApply}
\end{figure*}

\section{Performance in High Dimension}
The fundamental operation for the test is the computation of distances. However, as the dimension increases, the curse of dimensionality comes into play, pushing the average leaf community distance upon permuting for both dependent and independent features to be similar \cite{CoD}. Due to this, we can expect our test to break down in higher dimensions. This problem is particularly relevant in gene expression datasets, which typically have less than 500 observations (patients) but have thousands of features (genes). For instance, the gene expression dataset used by Singh et al. \cite{Singh} has 136 observations but 12600 features. Predicting disease is an important task, but so is understanding what genes are most useful for prediction. The first task would prefer the use of complex models such as the random forest, but genes are typically linked with one another \cite{corrFeaturesGenes}, making PFI unreliable. Furthermore, computationally intensive methods that leave out a feature and retrain the model have a prohibitive cost of computation.

\begin{figure}[t]
	\centering
	\captionsetup{justification=centering}
        \subfloat[Fifty observations\label{fig:twoCorrelatedCoD}]{\includegraphics[width=0.2\textwidth]{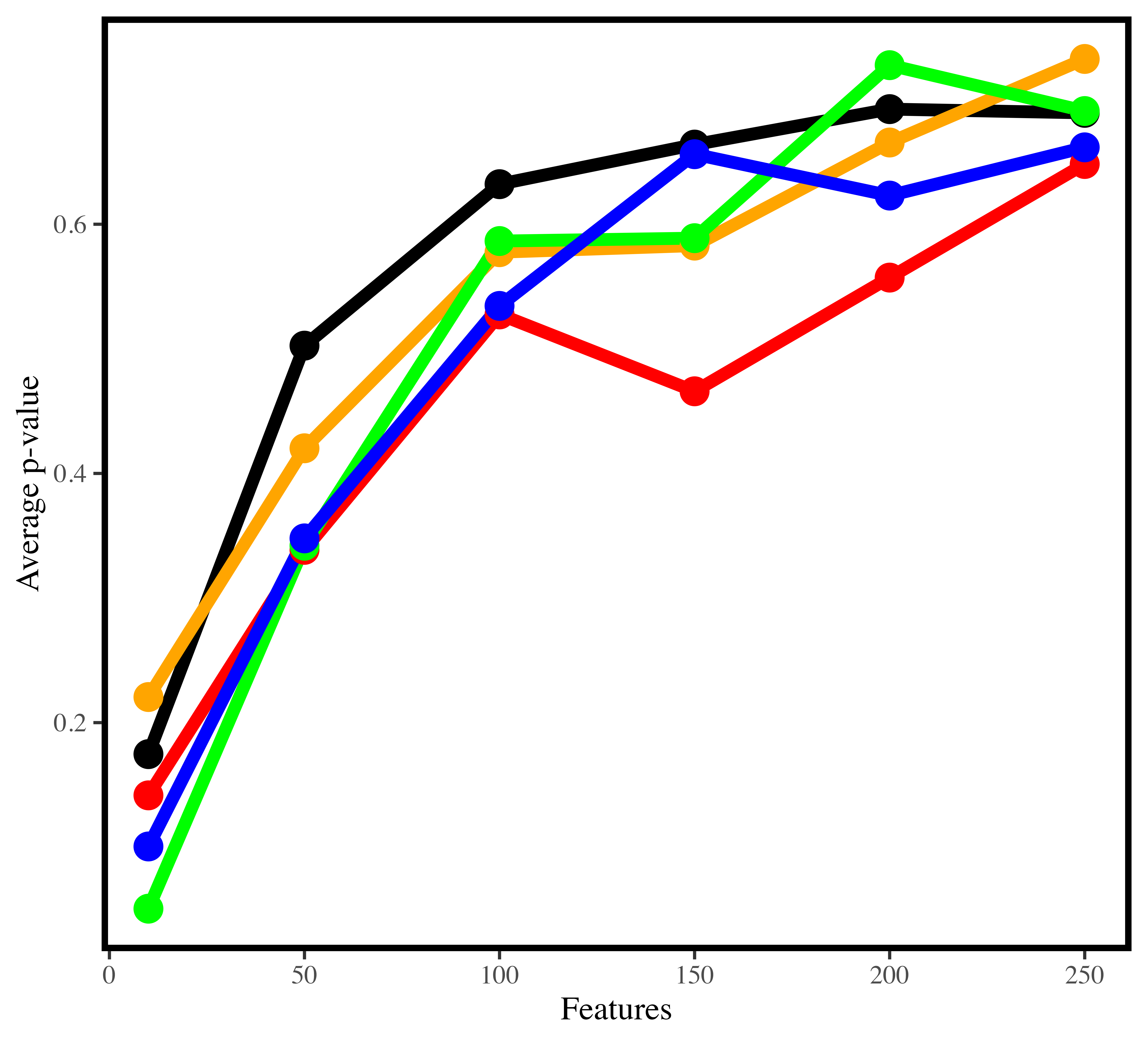}}\hfil
        \subfloat[100 observations\label{fig:justDependentCoD}]{\includegraphics[width=0.2\textwidth]{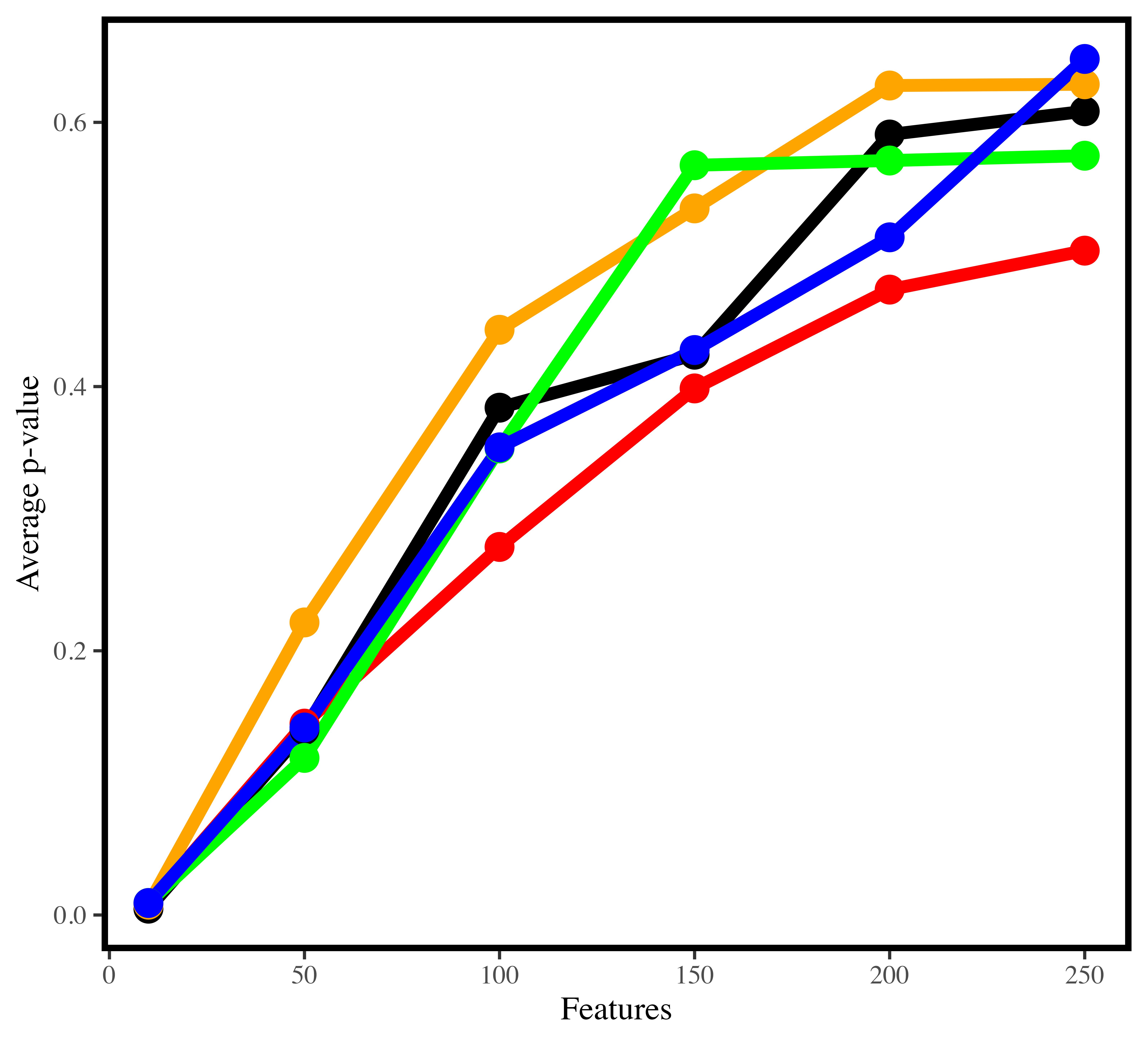}}\hfil
	\subfloat[150 observations\label{fig:blocksCoD}]{\includegraphics[width=0.2\textwidth]{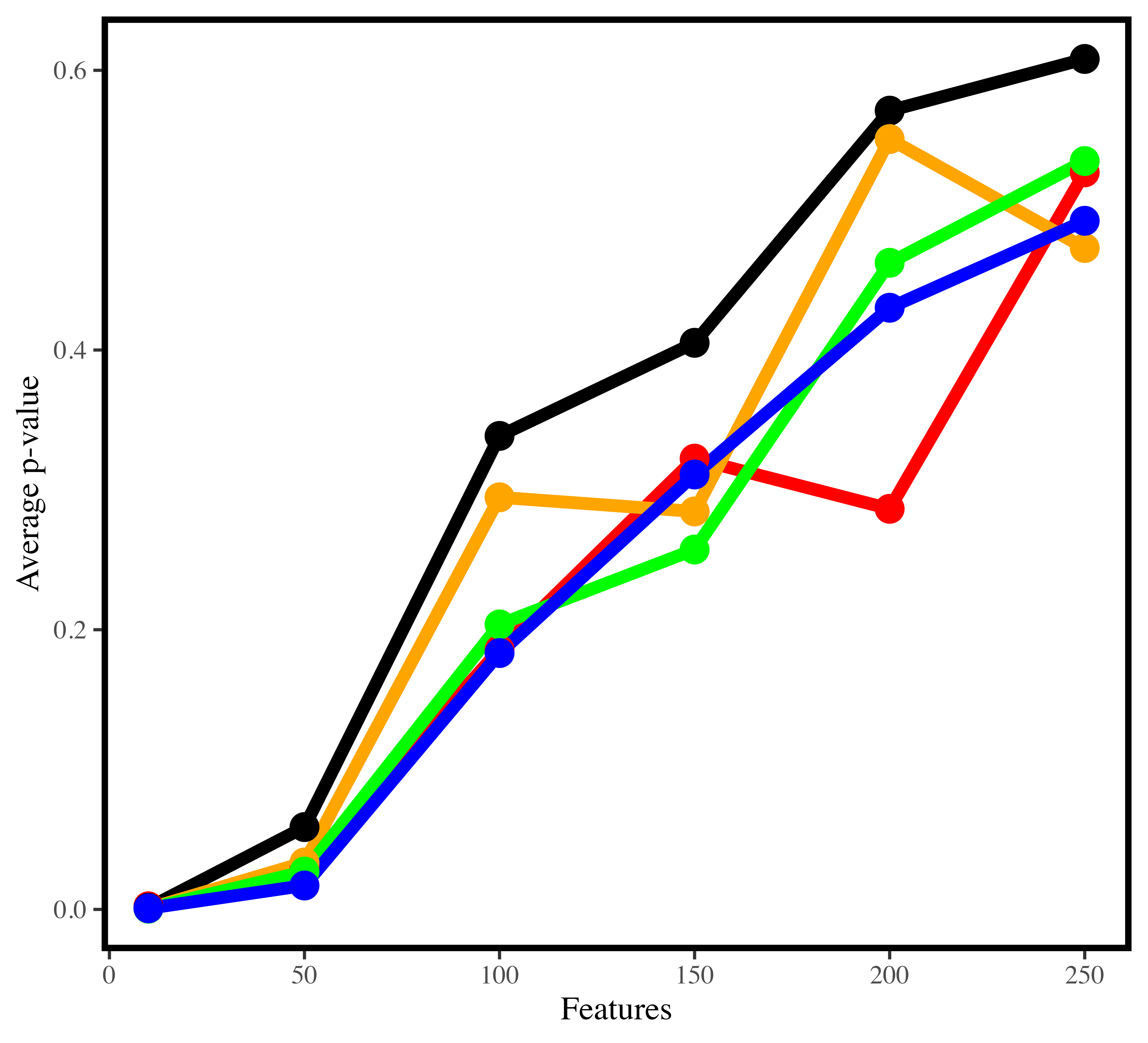}}\hfil
	\subfloat[Legend]{\includegraphics[width = 0.2\textwidth]{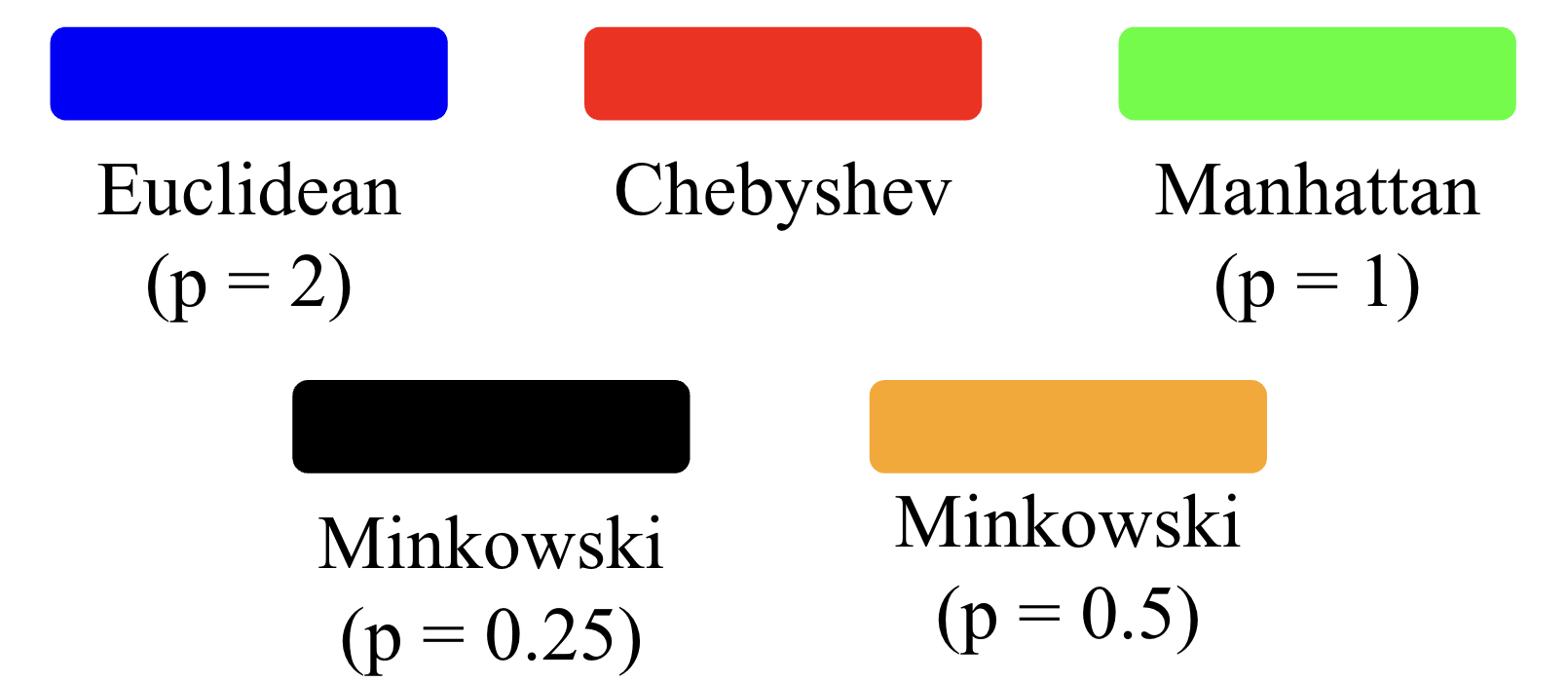}}
    \caption{Average p-value for a handful of metrics, across various dimensions of dataset.}
    \label{fig:CoD}
\end{figure}

We first demonstrate the curse of dimensionality at play for the test through simulation. To match the structure of the gene expression data, we will consider data sets with 50, 100, and 150 observations. Thankfully, the test begins to struggle far before the number of features grows into the thousands. We consider data sets with 10, 50, 100, 150, 200, and 250 features, and for each repeat the process of training the forest, permuting features, and computing distances as presented in the second chapter. The datasets will have blocks of ten correlated features. Thus far we have been working with the Euclidean metric, but here we will consider the broader Minkowski metric, of the form \[d(\mathbf{x},\mathbf{y}) = \left(\sum_{i=1}^n \vert x_i - y_i \vert^p \right)^{\frac{1}{p}}\] for $p>0$. We will try various values of $p$, including fractional values. For $p<1$, the triangle equality is violated, but it has been suggested that these fractional values of $p$ are useful for mitigating the curse of dimensionality \cite{fractionalMetrics}. We see that for all sample sizes, regardless of distance metric, as the number of features increases the p-values of the test begin to inflate as dependent features no longer register distances that are dissimilar from those that occur from permuting the independent baseline feature (Fig. \ref{fig:CoD}). The increase in sample size slightly improves the situation, but not nearly enough to save the test.

No modifications to the test will be made in order to allow it to be used with high dimension datasets. Instead, we will apply dimension reduction in order to reduce the number of features down to a level for which the test will not suffer from the curse of dimensionality. For the dimension reduction approach to be effective, the features must be uncorrelated or minimally correlated, and interpretable. After all, accurate importance scores for uninterpretable features do not do us much good. At first glance PCA looks enticing, given the uncorrelated nature of the components. However, each of the components are constructed as a linear combination of the original features. In practice, this results in dense, uninterpretable components \cite{myThesis}.

The sparse principal component analysis method (SPCA) recognizes PCA as a regression problem, and thus regularization penalties such as the ridge and lasso can be used to encourage sparsity \cite{SPCA}. In our case, we want to succinctly summarize each of the components, hence these penalties will be placed on the columns of the \emph{weight matrix}, which defines the components as linear combinations of the original features \cite{weightsLoadings}. This gives the following optimization problem:
\[\operatorname*{argmin}_{\mathbf{W}_R,\mathbf{P}_R} \lVert \mathbf{X} - \mathbf{XW}_R\mathbf{P}_R^T\rVert_F^2 + \lambda\sum_{r=1}^R \lVert \mathbf{w}_r \rVert_1 + \lambda_2\sum_{r=1}^R\lVert \mathbf{w}_r\rVert_2^2\]
subject to $\mathbf{P}_R\mathbf{P}_R^T = \mathbf{I}$. Note that the orthogonality constraint on the columns of $\mathbf{W}$ that is present in the PCA specification has been dropped. This simplifies the optimization problem but raises the concern that the components are now correlated. However, now that we're back to a smaller number of features, we can again apply our test to determine whether permuting these features does in fact cause extrapolation.

\subsection{Setup}

To evaluate this approach, we first apply it to a simulated data set. We work with a block structure with 150 total features and 25 observations in each block, with 100 observations in the training set and 50 in the test set. We then have six distinct blocks of features that we would like to summarize with dimension reduction, certainly manageable for our extrapolation test. With this, we move forward to applying our test to see if the importances for the new features are in fact reliable, using our test. After all, with the introduction of the sparsity to the regularization, there is no guarantee on the correlation of the components. For this, we generated the dataset outlined previously 25 times, and for each fit SPCA by gradually increasing $\lambda_1$ until each feature was only represented once in the components. With this sparse dimension reduction, we can transform our data and calculate importances and p-values as usual. Within each of the 25 iterations of generating the data, each feature is permuted 25 times for the calculation of importances and p-values. With only six features (other than the baseline), the curse of dimensionality is no longer a concern. 

Next we apply this approach to gene expression data. We work with the data used by Singh et al. in their study of gene expression data for the classification of patients as healthy or ill with prostate cancer \cite{Singh}. There are 12,600 genes to consider, so for computational reasons we will be following \cite{Witten} and only consider the 5\% of genes with the greatest variance, still leaving 630 genes. With only 102 observations in the training set and 34 in the test set, we should definitely be concerned about the curse of dimensionality. Seven components are all that is needed to explain most of the variance, and we will select the values for $\lambda_1$ and $\lambda_2$ such that the number of non-zero weights equals 630, the number of features in the dataset.

\subsection{Results}

\begin{figure*}[btp]
    \centering
    \hsize=\textwidth
    \captionsetup{justification=centering}
    \subfloat[PFI and cPFI for each sparse component.\label{fig:simImpsSPCA}]{\includegraphics[width=\columnwidth]{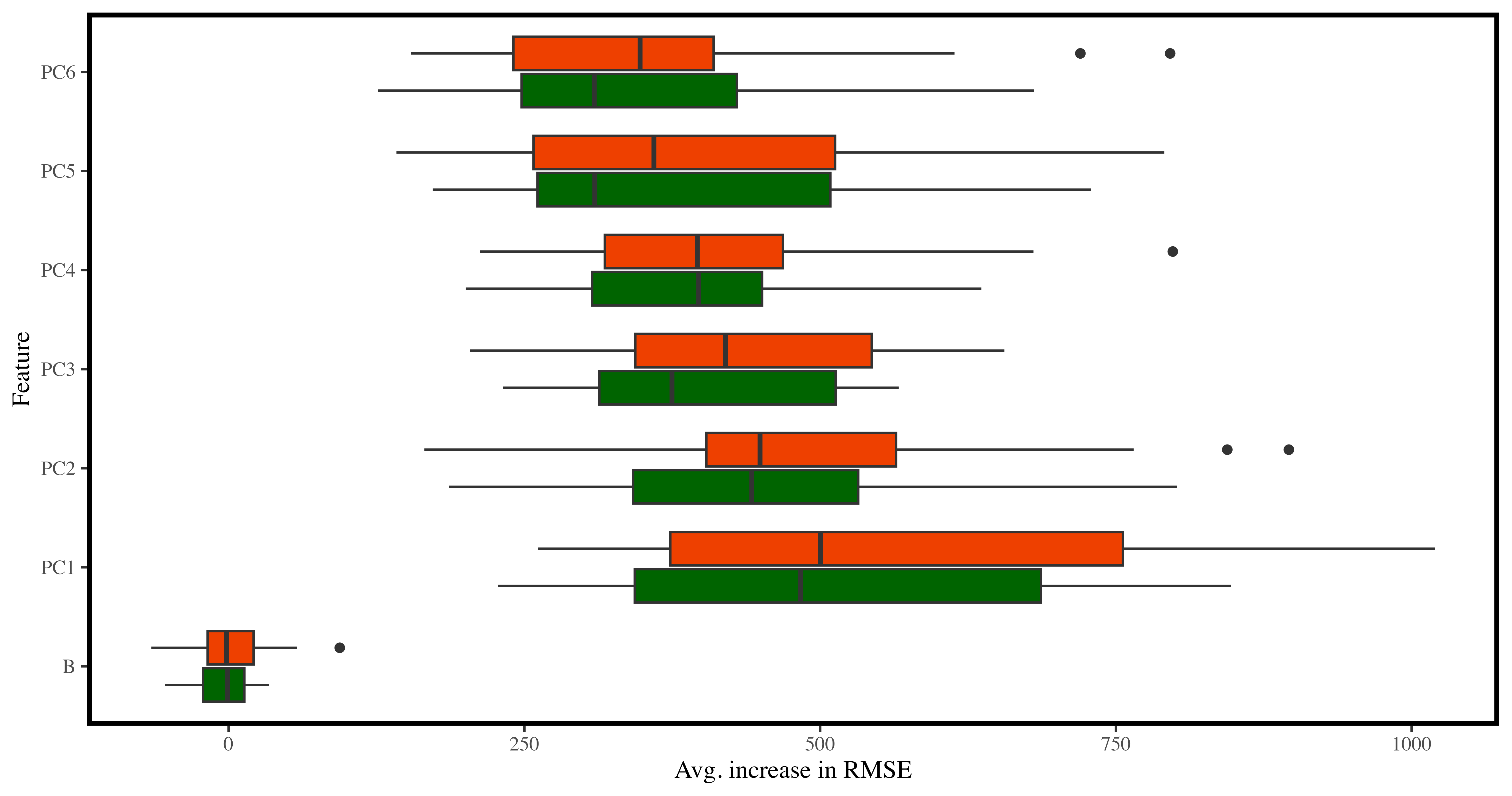}}\hfil
    \subfloat[Boxplots of p-values for each sparse component. \label{fig:simPValsSPCA}]{\includegraphics[width=\columnwidth]{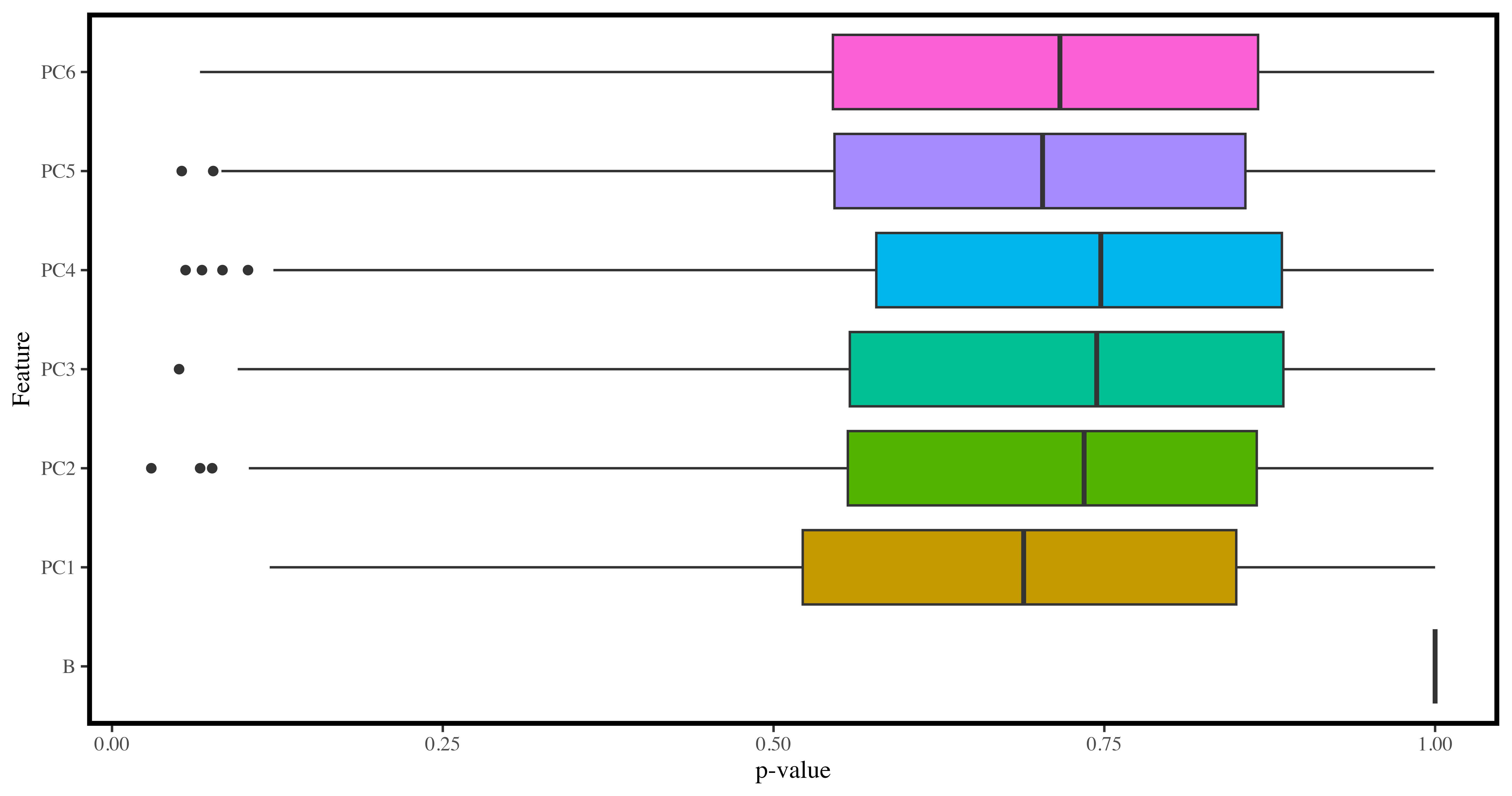}}\hfil
    \caption{Feature importance/testing for the simulated dataset. The overlap in importances indicates little to no extrapolation, which is supported by the boxplots of p-values, which all have three quartiles above 0.5.}
    \label{fig:simResultsSPCA}
\end{figure*}

We can see that all components have median p-values around 0.7, with none of the features having their lower quartile of p-values across 625 permutations below 0.5, suggesting that we can work under the belief that our model does not extrapolate upon permuting the features, thus yielding reliable PFI scores. This is confirmed when we compare PFI and cPFI importances, which overlap considerably for all components (Fig. \ref{fig:simPValsSPCA}). Since the SPCA method is a modification of PCA -- which guarantees correlations of zero -- we would expect that the components are not strongly correlated. However, by adding the sparsity penalties we lose the \emph{guarantee} of uncorrelated components, but the test can be used to evaluate whether we have created components for which we can trust their PFI importance scores.

\begin{figure*}[tb]
	\centering
	\hsize=\textwidth
	\captionsetup{justification=centering}
	\subfloat[Average PFI and cPFI across permutations.\label{fig:SinghImps}]{\includegraphics[width=0.33\textwidth]{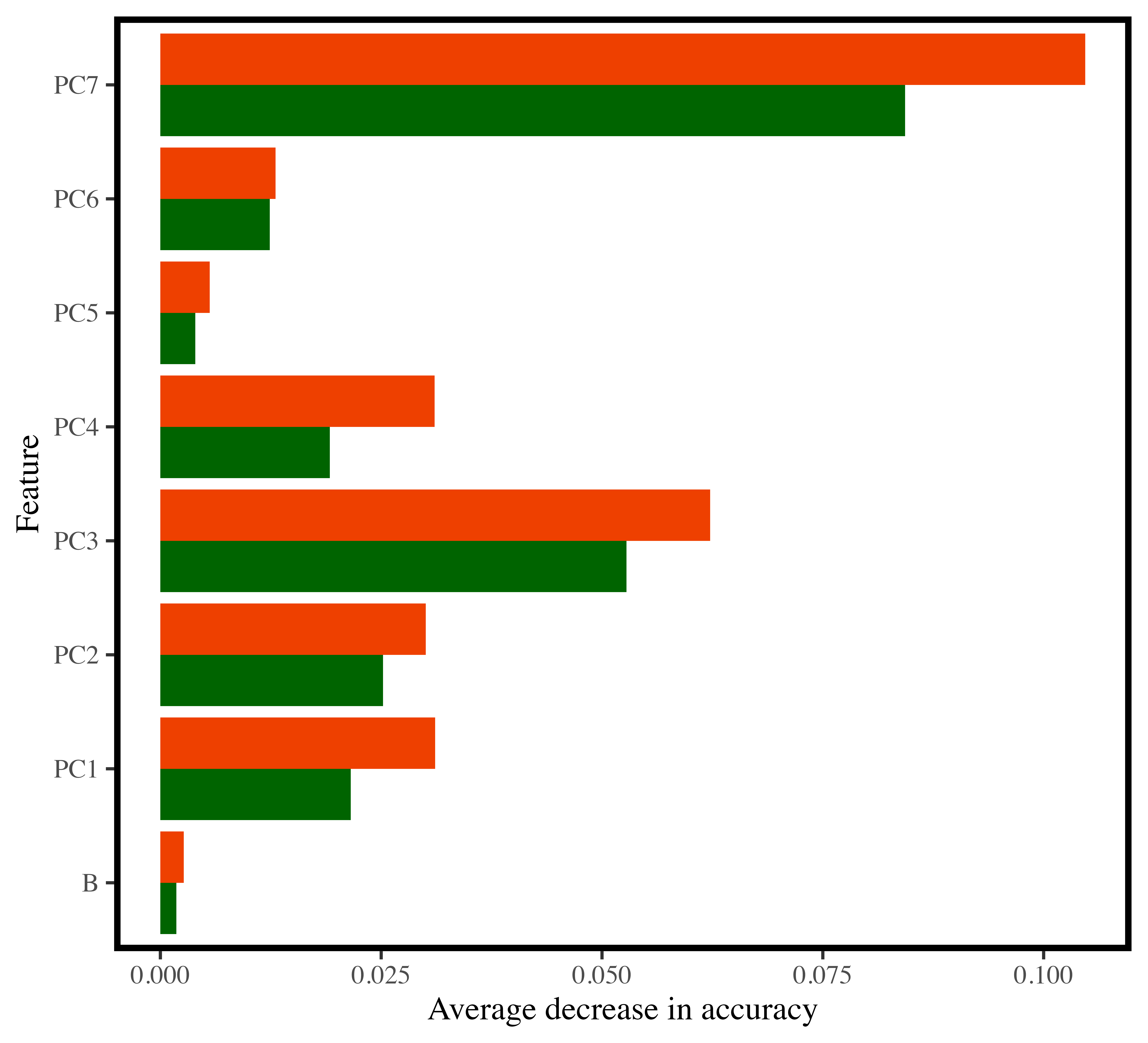}}\hfil
        \subfloat[Boxplots of p-values from 25 permutations.\label{fig:SinghPVals}]{\includegraphics[width=0.33\textwidth]{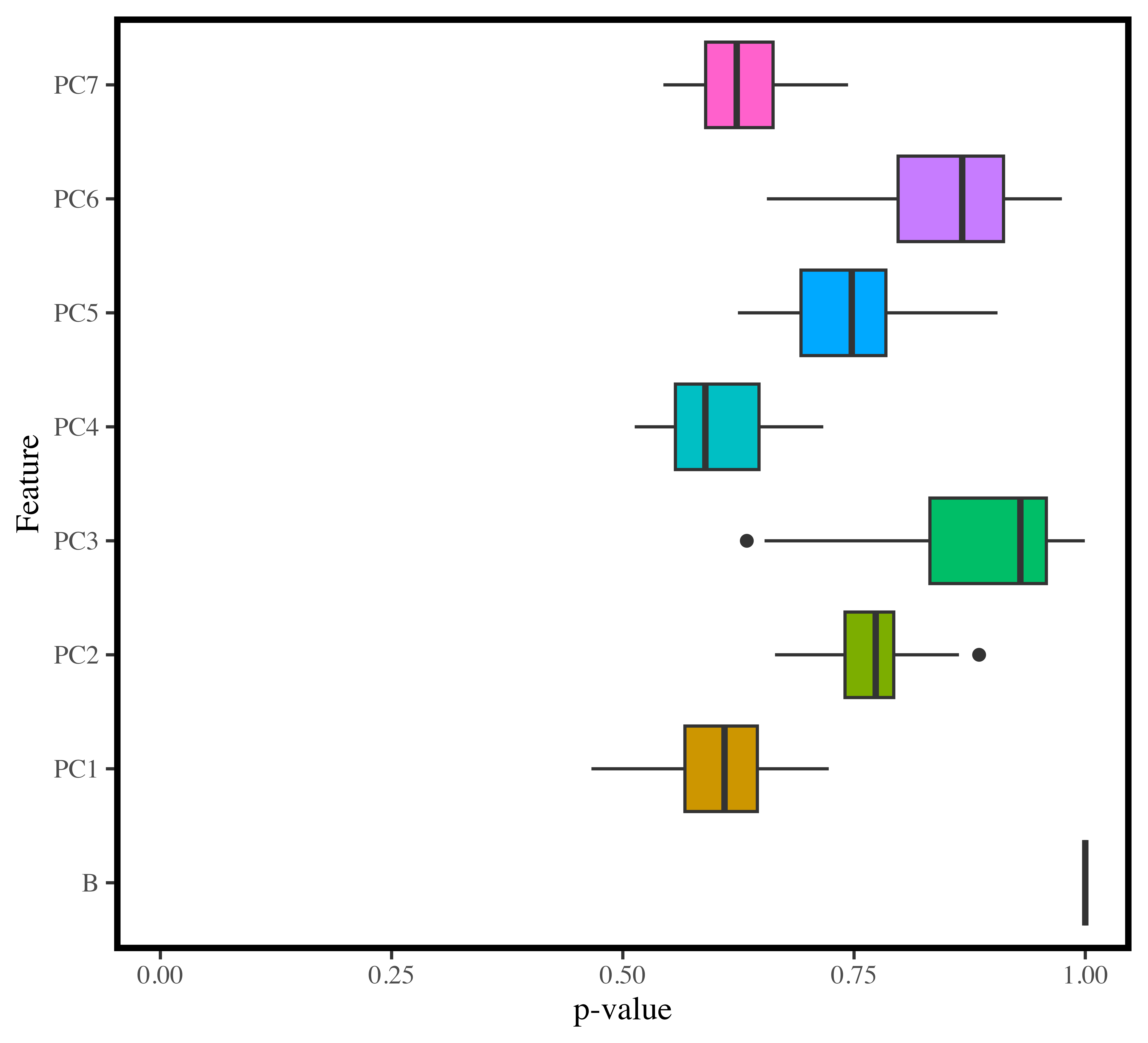}}\hfil
	\subfloat[Correlation heat map for the dimension-reduced test set.\label{fig:SinghCorr}]{\includegraphics[width=0.33\textwidth]{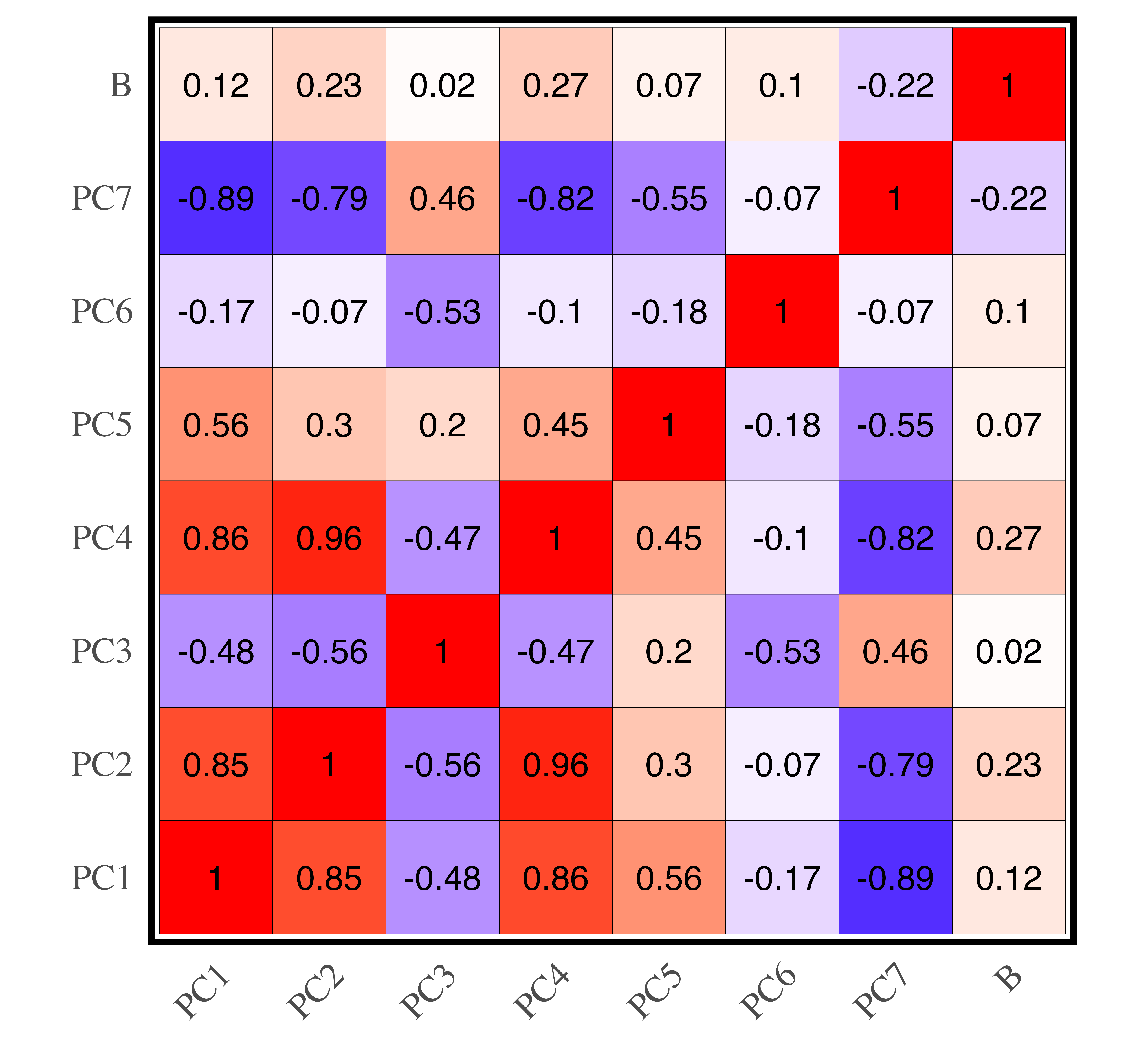}}
    \caption{Feature importance/testing for the prostate cancer gene expression data. Components two, four, and seven do appear to extrapolate to the point of biasing importance scores, while the others do not.}
    \label{fig:SinghResults}
\end{figure*}

In the gene expression data, after applying SPCA we see that for some components cPFI and PFI are close, but further consideration is warranted for others (Fig. \ref{fig:SinghImps}). Components PC6 and PC3 -- with boxplots of p-values the furthest right -- have PFI and cPFI that are very close. On the other hand, components PC7, PC4, and PC2 show the greatest difference in cPFI and PFI scores, along with boxplots of p-values that are the farthest left (Fig. \ref{fig:SinghPVals}). With only 34 observations with which to compute p-values, it is unlikely to produce strongly indicative results, but the separation in p-values for some features could be taken as an indicator that more computationally intensive feature importance scores should be considered. Looking at a correlation heat map of the components (Fig. \ref{fig:SinghCorr}), we see that in fact components two, four, and seven are more strongly correlated with the other components than those with the higher p-values. With just this correlation heat map, it would be difficult to determine that PFI would not be a reliable feature importance method, but in combination with our test a practitioner is empowered to make a more informed decision.

\section{Conclusion}
It is well understood that popular efficient feature importance methods can be unreliable, prioritizing certain types of features or giving preference to dependent features. In this work we propose a testing framework to evaluate the extent to which the model extrapolates when permuting the feature, and with this the reliability of the importance scores can be evaluated. We demonstrated the applicability to both simulated and real low-dimension data. However, for higher dimensions, dimension reduction approaches are necessary, and the reduction must produce interpretable features. Through the application of SPCA interpretable components can be extracted, yielding a new low-dimension problem that can be tackled. We leave an analysis of the sensitivity of the method to the chosen distance metric as future work.

\bibliographystyle{named}
\bibliography{ijcai25}

\begin{thebibliography}{}

\bibitem[\protect\citeauthoryear{Aeberhard and Forina}{1992}]{Wine}
Stefan Aeberhard and M.~Forina.
\newblock {Wine}.
\newblock UCI Machine Learning Repository, 1992.
\newblock {DOI}: https://doi.org/10.24432/C5PC7J.

\bibitem[\protect\citeauthoryear{Aggarwal \bgroup \em et al.\egroup
  }{2001}]{fractionalMetrics}
Charu~C. Aggarwal, Alexander Hinneburg, and Daniel~A. Keim.
\newblock On the surprising behavior of distance metrics in high dimensional
  space.
\newblock In Jan Van~den Bussche and Victor Vianu, editors, {\em Database
  Theory --- ICDT 2001}, pages 420--434, Berlin, Heidelberg, 2001. Springer
  Berlin Heidelberg.

\bibitem[\protect\citeauthoryear{Ahneman \bgroup \em et al.\egroup
  }{2018}]{annePaper}
Derek~T. Ahneman, Jesús~G. Estrada, Shishi Lin, Spencer~D. Dreher, and
  Abigail~G. Doyle.
\newblock Predicting reaction performance in c–n cross-coupling using machine
  learning.
\newblock {\em Science}, 360(6385):186--190, 2018.

\bibitem[\protect\citeauthoryear{Alex~Goldstein and
  Pitkin}{2015}]{individualConditionalExpectationPlot}
Justin~Bleich Alex~Goldstein, Adam~Kapelner and Emil Pitkin.
\newblock Peeking inside the black box: Visualizing statistical learning with
  plots of individual conditional expectation.
\newblock {\em Journal of Computational and Graphical Statistics},
  24(1):44--65, 2015.

\bibitem[\protect\citeauthoryear{Almog and Kalech}{2023}]{conceptDrift}
Shaked Almog and Meir Kalech.
\newblock Diagnosis for post concept drift decision trees repair.
\newblock In {\em Proceedings of the 20th International Conference on
  Principles of Knowledge Representation and Reasoning}, KR '23, 2023.

\bibitem[\protect\citeauthoryear{Archer and Kimes}{2008}]{ARCHER}
Kellie~J. Archer and Ryan~V. Kimes.
\newblock Empirical characterization of random forest variable importance
  measures.
\newblock {\em Computational Statistics \& Data Analysis}, 52(4):2249--2260,
  2008.

\bibitem[\protect\citeauthoryear{Beyer \bgroup \em et al.\egroup }{1999}]{CoD}
Kevin Beyer, Jonathan Goldstein, Raghu Ramakrishnan, and Uri Shaft.
\newblock When is ``nearest neighbor'' meaningful?
\newblock In Catriel Beeri and Peter Buneman, editors, {\em Proceedings of the
  7th International Conference on Database Theory, Jerusalem, Israel}, Lecture
  Notes in Computer Science, pages 217--235. Springer Berlin Heidelberg, 1999.

\bibitem[\protect\citeauthoryear{Brazma and Vilo}{2000}]{corrFeaturesGenes}
Alvis Brazma and Jaak Vilo.
\newblock Gene expression data analysis.
\newblock {\em FEBS Letters}, 480(1):17--24, 2000.

\bibitem[\protect\citeauthoryear{Breiman \bgroup \em et al.\egroup
  }{1984}]{CART}
L.~Breiman, J.~Friedman, C.J. Stone, and R.A. Olshen.
\newblock {\em Classification and Regression Trees}.
\newblock Taylor \& Francis, 1984.

\bibitem[\protect\citeauthoryear{Breiman}{2001}]{randomForest}
Leo Breiman.
\newblock Random forests.
\newblock {\em Machine Learning}, 45(1):5--32, October 2001.

\bibitem[\protect\citeauthoryear{Debeer and Strobl}{2020}]{revisitCPFI}
Dries Debeer and Carolin Strobl.
\newblock Conditional permutation importance revisited.
\newblock {\em BMC Bioinformatics}, 21(1):307, 2020.

\bibitem[\protect\citeauthoryear{Debeer \bgroup \em et al.\egroup
  }{2021}]{cPFI}
Dries Debeer, Torsten Hothorn, and Carolin Strobl.
\newblock {\em permimp: Conditional Permutation Importance}, 2021.
\newblock R package version 1.0-2.

\bibitem[\protect\citeauthoryear{Foote}{2025}]{myThesis}
Aaron Foote.
\newblock Diagnosing biased feature importance scores in tree-based models.
\newblock Master's thesis, Wesleyan University, 2025.

\bibitem[\protect\citeauthoryear{Friedman}{2001}]{partialDependencePlot}
Jerome~H. Friedman.
\newblock {Greedy function approximation: A gradient boosting machine.}
\newblock {\em The Annals of Statistics}, 29(5):1189 -- 1232, 2001.

\bibitem[\protect\citeauthoryear{Garrett and
  Rudin}{2023}]{needInterpretability1}
Brandon~L. Garrett and Cynthia Rudin.
\newblock Interpretable algorithmic forensics.
\newblock {\em Proceedings of the National Academy of Sciences}, 120(41), 2023.

\bibitem[\protect\citeauthoryear{Grinsztajn \bgroup \em et al.\egroup
  }{2022}]{forestOverDL2}
L\'{e}o Grinsztajn, Edouard Oyallon, and Ga\"{e}l Varoquaux.
\newblock Why do tree-based models still outperform deep learning on typical
  tabular data?
\newblock In {\em Proceedings of the 36th International Conference on Neural
  Information Processing Systems}, NIPS '22, pages 507--520, Red Hook, NY, USA,
  2022. Curran Associates Inc.

\bibitem[\protect\citeauthoryear{Hooker \bgroup \em et al.\egroup
  }{2021}]{noFreeVariableImportance}
Giles Hooker, Lucas Mentch, and Siyu Zhou.
\newblock Unrestricted permutation forces extrapolation: variable importance
  requires at least one more model, or there is no free variable importance.
\newblock {\em Statistics and Computing}, 31(6):82, 2021.

\bibitem[\protect\citeauthoryear{Hooker}{2007}]{functionalANOVA}
Giles Hooker.
\newblock Generalized functional anova diagnostics for high-dimensional
  functions of dependent variables.
\newblock {\em Journal of Computational and Graphical Statistics},
  16(3):709--732, 2007.

\bibitem[\protect\citeauthoryear{Jacobs \bgroup \em et al.\egroup
  }{2021}]{medicineML}
Maia Jacobs, Melanie~F. Pradier, Thomas~H. McCoy, Roy~H. Perlis, Finale
  Doshi-Velez, and Krzysztof~Z. Gajos.
\newblock How machine-learning recommendations influence clinician treatment
  selections: the example of antidepressant selection.
\newblock {\em Translational Psychiatry}, 11(1):108, 2021.

\bibitem[\protect\citeauthoryear{Lei \bgroup \em et al.\egroup }{2018}]{loco}
Jing Lei, Max G'Sell, Alessandro Rinaldo, Ryan~J. Tibshirani, and Larry
  Wasserman.
\newblock Distribution-free predictive inference for regression.
\newblock {\em Journal of the American Statistical Association},
  113(523):1094--1111, 07 2018.

\bibitem[\protect\citeauthoryear{Liaw and Wiener}{2002}]{randomForestPackageR}
Andy Liaw and Matthew Wiener.
\newblock Classification and regression by randomforest.
\newblock {\em R News}, 2(3):18--22, 2002.

\bibitem[\protect\citeauthoryear{Liu \bgroup \em et al.\egroup
  }{2019}]{mlProfessional}
Xiaoxuan Liu, Livia Faes, Aditya~U Kale, Siegfried~K Wagner, Dun~Jack Fu, Alice
  Bruynseels, Thushika Mahendiran, Gabriella Moraes, Mohith Shamdas, Christoph
  Kern, Joseph~R Ledsam, Martin~K Schmid, Konstantinos Balaskas, Eric~J Topol,
  Lucas~M Bachmann, Pearse~A Keane, and Alastair~K Denniston.
\newblock A comparison of deep learning performance against health-care
  professionals in detecting diseases from medical imaging: a systematic review
  and meta-analysis.
\newblock {\em The Lancet Digital Health}, 1(6):e271--e297, 2019.

\bibitem[\protect\citeauthoryear{Lundberg and Lee}{2017}]{SHAP}
Scott~M. Lundberg and Su-In Lee.
\newblock A unified approach to interpreting model predictions.
\newblock In {\em Proceedings of the 31st International Conference on Neural
  Information Processing Systems}, NIPS'17, page 4768–4777, Red Hook, NY,
  USA, 2017. Curran Associates Inc.

\bibitem[\protect\citeauthoryear{Mentch and Hooker}{2016}]{permRelearn}
Lucas Mentch and Giles Hooker.
\newblock Quantifying uncertainty in random forests via confidence intervals
  and hypothesis tests.
\newblock {\em Journal of Machine Learning Research}, 17(26):1--41, 2016.

\bibitem[\protect\citeauthoryear{Nicodemus \bgroup \em et al.\egroup
  }{2010}]{conditionalPermutation}
Kristin~K. Nicodemus, James~D. Malley, Carolin Strobl, and Andreas Ziegler.
\newblock The behaviour of random forest permutation-based variable importance
  measures under predictor correlation.
\newblock {\em BMC Bioinformatics}, 11(1):110, 2010.

\bibitem[\protect\citeauthoryear{Park \bgroup \em et al.\egroup
  }{2024}]{weightsLoadings}
S.~Park, E.~Ceulemans, and K.~Van~Deun.
\newblock A critical assessment of sparse pca (research): why (one should
  acknowledge that) weights are not loadings.
\newblock {\em Behavior Research Methods}, 56(3):1413--1432, 2024.

\bibitem[\protect\citeauthoryear{Probst \bgroup \em et al.\egroup
  }{2019}]{minimalTuning}
Philipp Probst, Marvin~N. Wright, and Anne-Laure Boulesteix.
\newblock Hyperparameters and tuning strategies for random forest.
\newblock {\em WIREs Data Mining and Knowledge Discovery}, 9(3):e1301, 2019.

\bibitem[\protect\citeauthoryear{Shwartz-Ziv and Armon}{2022}]{forestOverDL1}
Ravid Shwartz-Ziv and Amitai Armon.
\newblock Tabular data: Deep learning is not all you need.
\newblock {\em Information Fusion}, 81:84--90, 2022.

\bibitem[\protect\citeauthoryear{Singh \bgroup \em et al.\egroup
  }{2002}]{Singh}
Dinesh Singh, Phillip~G Febbo, Kenneth Ross, Donald~G Jackson, Judith Manola,
  Christine Ladd, Pablo Tamayo, Andrew~A Renshaw, Anthony~V D'Amico, Jerome~P
  Richie, Eric~S Lander, Massimo Loda, Philip~W Kantoff, Todd~R Golub, and
  William~R Sellers.
\newblock Gene expression correlates of clinical prostate cancer behavior.
\newblock {\em Cancer Cell}, 1(2):203--209, mar 2002.

\bibitem[\protect\citeauthoryear{Strobl \bgroup \em et al.\egroup
  }{2007}]{giniBiased}
Carolin Strobl, Anne-Laure Boulesteix, Achim Zeileis, and Torsten Hothorn.
\newblock Bias in random forest variable importance measures: Illustrations,
  sources and a solution.
\newblock {\em BMC Bioinformatics}, 8(1):25, 2007.

\bibitem[\protect\citeauthoryear{Strobl \bgroup \em et al.\egroup
  }{2008}]{strobl3}
Carolin Strobl, Anne-Laure Boulesteix, Thomas Kneib, Thomas Augustin, and Achim
  Zeileis.
\newblock Conditional variable importance for random forests.
\newblock {\em BMC Bioinformatics}, 9(1):307, 2008.

\bibitem[\protect\citeauthoryear{Witten \bgroup \em et al.\egroup
  }{2009}]{Witten}
Daniela~M Witten, Robert Tibshirani, and Trevor Hastie.
\newblock A penalized matrix decomposition, with applications to sparse
  principal components and canonical correlation analysis.
\newblock {\em Biostatistics}, 10(3):515--534, apr 2009.

\bibitem[\protect\citeauthoryear{Zou \bgroup \em et al.\egroup }{2006}]{SPCA}
Hui Zou, Trevor Hastie, and Robert Tibshirani.
\newblock Sparse principal component analysis.
\newblock {\em Journal of Computational and Graphical Statistics},
  15(2):265--286, 2006.

\end{thebibliography}

\end{document}